\documentclass{article}

\usepackage{microtype}
\usepackage{graphicx}
\usepackage{subcaption}
\usepackage{booktabs} 

\usepackage{hyperref}

\usepackage[preprint]{icml2026}

\usepackage{amsmath}
\usepackage{amssymb}
\usepackage{mathtools}
\usepackage{amsthm}
\usepackage[inline]{enumitem}
\usepackage{multirow}

\usepackage[capitalize,noabbrev]{cleveref}

\theoremstyle{plain}

\theoremstyle{definition}

\theoremstyle{remark}

\usepackage[disable,textsize=tiny]{todonotes}

\icmltitlerunning{Scaling Next-Brain-Token Prediction for MEG}

\begin{document}

\twocolumn[
  \icmltitle{Scaling Next-Brain-Token Prediction for MEG}



  \icmlsetsymbol{equal}{*}

  \begin{icmlauthorlist}
    \icmlauthor{Richard Csaky}{for}

  \end{icmlauthorlist}

  \icmlaffiliation{for}{Foresight Institute}

  \icmlcorrespondingauthor{Richard Csaky}{richard.csaky@gmail.com}

  \icmlkeywords{Magnetoencephalography, Time series, Autoregressive modeling, Generative modeling, GPT, Tokenization, Conditional generation, multi-subject modeling}

  \vskip 0.3in
]



\printAffiliationsAndNotice{}  

\begin{abstract}
We present a large autoregressive model for source-space MEG that scales next-token prediction to long context across datasets and scanners: handling a corpus of over 500 hours and thousands of sessions across the three largest MEG datasets. A modified SEANet-style vector-quantizer reduces multichannel MEG into a flattened token stream on which we train a Qwen2.5-VL backbone from scratch to predict the next brain token and to recursively generate minutes of MEG from up to a minute of context. To evaluate long-horizon generation, we introduce two task-matched tests: (i) on-manifold stability via generated-only drift compared to the time-resolved distribution of real sliding windows, and (ii) conditional specificity via correct context versus prompt-swap controls using a neurophysiologically grounded metric set. We train on CamCAN and Omega and run all analyses on held-out MOUS, establishing cross-dataset generalization. Across metrics, generations remain relatively stable over long rollouts and are closer to the correct continuation than swapped controls.
 Code available at: \url{https://github.com/ricsinaruto/brain-gen}.
\end{abstract}

\section{Introduction}

Predicting the future---the next sensory input, the next state of the world, the next action---is central to both natural and artificial intelligence. In neuroscience, predictive coding and the free-energy principle frame perception and cognition as continual prediction and correction through prediction errors \citep{rao1999predictive,friston2010free}. In machine learning, recent progress has revealed a similar unification across domains: many successful systems can be viewed as solving one and the same causal modeling problem, \emph{given the past, what happens next?}

Across domains, the most successful instantiation of this recipe has been scaling flexible sequence models \citep{Vaswani:2017}. Whether through diffusion, normalizing flows, or discrete tokens \citep{ho2020ddpm,kingma2018glow,van2017neural}, large Transformer models can predict the next word, the next image patch, the next video frame or audio chunk, and the next action, increasingly exhibiting representations that behave like implicit world models \citep{vafa2024evaluating}. This raises a natural question: if implicit models of the world can emerge from predicting observations of the world (video, audio) or of human behavior (language), what kind of models might emerge from predicting the process that produces intelligent behavior itself: brain activity?

Brain recordings provide a privileged view into the internal dynamics that mediate perception, cognition, and action. In the \emph{learning using privileged information} (LUPI) framework, extra signals available at training time can improve generalization by exposing latent variables that are causally upstream of the observed outputs \citep{vapnik2009privileged}. Neural signals can act as such a privileged training signal: brain-based objectives can regularize vision models toward robust representations \citep{li2019brainsregularize}, and ``brain-tuning'' speech language models on fMRI can induce brain-relevant semantics and improve downstream performance \citep{moussa2024braintuning}. We therefore argue that a powerful \emph{generative} model of brain signals will have to internalize reusable structure about these dynamics, and indirectly, about the world models and future-state predictions the brain itself implements. Such a model could be useful both scientifically (simulation, data augmentation, probing) and as a source of grounding or ``privileged teaching'' for AI systems trained primarily on observations of behavior. This motivates our quest here as a first step in this direction: \emph{apply the causal prediction and scaling paradigm to brain data}.

We focus on magnetoencephalography (MEG), a unique and comparatively under-explored non-invasive modality which provides millisecond temporal resolution and high information density relative to EEG. MEG differs sharply from standard generative domains: it is a multi-channel time series of continuous values with low SNR and weak human interpretability, making both modeling and evaluation challenging. Still, the same scaling paradigm that has worked for language, audio, and video should apply to MEG, provided we can represent it as a token sequence that a modern generative backbone can model efficiently. Here, we use \emph{scaling} primarily to mean scaling in data: making a single model work across hundreds of hours and thousands of sessions drawn from multiple datasets and scanners, and then evaluating out-of-distribution on a fully held-out dataset. This is hard: MEG variability across sessions (subjects, tasks, hardware) is large. In exploratory baselines, several channel-mixing sequence models that worked on a handful of sessions collapsed when trained even on a full single-dataset corpus, suggesting that implicit robustness to variability is a key bottleneck.

We are especially interested in pushing this recipe in terms of context length and \emph{conditional specificity}. We do not only want to generate plausible brain activity; we want to generate brain activity that is specific to the context. Just as an LLM or VLM should respond differently to different prompts, a large brain model should produce \textit{long-range on-manifold signals that are conditionally specific to the session, subject, and task implied purely by the prompt input} (without auxiliary embeddings or metadata). This framing also makes brain models compatible with ``token-stream'' multimodal backbones: in the long run, brain tokens could be interleaved with language, vision, and action tokens in a single causal sequence.

Token-stream modeling is becoming a dominant design pattern in multimodal generative systems: high-bandwidth modalities are first mapped to compact sequences of tokens, and a single decoder-only backbone is trained over interleaved multimodal sequences. Emu3 and Emu3.5 show that a native multimodal Transformer trained solely with next-token prediction over unified vision-language tokens can support both perception and high-fidelity generation, including video synthesis \citep{wang2024emu3,cui2025emu35}. Qwen2.5-VL and Qwen3-VL extend this token-stream interface to high-resolution inputs and long-video understanding \citep{bai2025qwen25vl,qwen2024qwen25}.

To summarize, a good generative MEG model should have:
\begin{enumerate*}
    \item Token-based AR without auxiliary information.
    \item Conditional specificity to the input prompt.
    \item The ability to ingest long context.
    \item Stable and on-manifold long-horizon generation.
    \item An efficient and scalable architecture.
\end{enumerate*}

To achieve these goals, we propose applying the causal prediction paradigm to MEG by improving the BrainOmni tokenizer \citep{xiao2025brainomni} and training a Qwen-2.5-VL-style decoder-only backbone from scratch for next-brain-token prediction, without auxiliary task/dataset information. We scale this to the three largest publicly accessible MEG datasets: CamCAN, Omega, and MOUS \citep{taylor2017camcan,niso2016omega,schoffelen2019mous}, with a combined size of over 500 hours across rest and many diverse tasks. We train on CamCAN and OMEGA and report all results on MOUS, a fully held-out dataset with out-of-distribution tasks.

To address the signal interpretation issue, we propose an extensive evaluation framework comparing neurophysiologically grounded metrics across multiple minutes of free-running recursive generation. Our protocol is designed to evaluate long-horizon on-manifold stability, conditional specificity (via prompt-swap controls), and variability calibration (via a task-matched real-real baseline). Since MEG signals are not directly interpretable to humans, evaluation frameworks that mimic long-range stress tests used for LLMs are especially important.

\textbf{Contributions.}
\begin{itemize}
    \item \texttt{BrainTokMix}, a causal channel-mixing RVQ tokenizer for source-space MEG.
    \item \texttt{FlatGPT}, a decoder-only Transformer trained from scratch on \texttt{BrainTokMix} tokens using standard next-token cross-entropy, enabling multi-minute prompt-and-generate MEG rollouts without auxiliary metadata.
    \item A \emph{cross-dataset} evaluation protocol that stress-tests long-horizon stability and prompt dependence using neurophysiological metrics and prompt-swap controls.
\end{itemize}

\section{Related Work}

Neural foundation modeling has rapidly adopted self-supervised pretraining and discrete tokenization for heterogeneous EEG/MEG.
LaBraM pretrains Transformers over quantized EEG patches via masked prediction \citep{jiang2024labram};
BrainOmni introduces a sensor-aware tokenizer and unified EEG/MEG pretraining \citep{xiao2025brainomni};
and NeuroRVQ studies multi-scale RVQ codebooks for MEG tokenization \citep{barmpas2025neurorvq}. Generative models for electrophysiology include autoregressive code models (e.g., MEG-GPT, which focuses on sub-second contexts) and diffusion-style approaches \citep{lim2024eegtrans,huang2025meggpt}. Compared to these efforts, \texttt{FlatGPT} emphasizes (i) purely next-token objective
over discrete MEG tokens through an efficient and scalable
paradigm (ii) long-context conditioning through prompting
rather than labels, and (iii) stress-testing open-loop generations for stability and specificity using an out-of-distribution evaluation.

Outside neuroscience, time-series foundation models and discretization approaches provide scalable forecasting recipes \citep{tsfm_das2024timesfm,ansari2024chronos}, and token-stream multimodal decoders in vision and audio motivate treating high-bandwidth signals as tokens for a single causal backbone \citep{defossez2022hifi,wang2024emu3,bai2025qwen25vl,agarwal2025cosmos}. Due to space constraints we provide an extended related-work discussion in Appendix~\cref{app:related_work}.

\section{Methods}

MEG poses an unusual combination of challenges for modern generative modeling: high sampling rate (here 100\,Hz), long temporal horizons (tens of seconds to minutes), and multichannel structure ($C\!=\!68$ source-space regions in our main setup).

The following inductive biases summarize the constraints that guided our method design:
\begin{enumerate*}\itemsep0.25em
    \item Minute-scale context.
    \item Spatiotemporal tokens: A token should represent a temporally and spatially reduced patch of MEG.
    \item Flatten into a single sequence: Serialize temporal and spatial axes into one token stream to enable full attention mixing under a standard causal mask.
    \item Prompt-only: Do not rely on specialized conditioning embeddings.
    \item Pure next-token objective.
\end{enumerate*}

We therefore build \texttt{FlatGPT} around a simple but scalable decomposition that mirrors frontier LLM/VLM pipelines:
(i) learn a causal discrete tokenizer that compresses multichannel MEG into a grid of discrete code indices,
and (ii) train a decoder-only Transformer with only teacher-forced next-token prediction (cross-entropy) in token space.
This choice is motivated by both scalability and interoperability: once MEG becomes a token stream, we can directly leverage the same decoder-only architectures used for language, audio, and video token streams, and later even interleave with these modalities in a unified token sequence. Appendix \cref{tab:positioning} summarizes where our proposed method sits compared to prior art.

\subsection{Problem formulation}
A preprocessed MEG recording is a multichannel time series $x \in \mathbb{R}^{C \times T}$ here sampled at $f_s=100$\,Hz.
Given a context $x_{:t}$, we aim to model the conditional distribution over future activity,
\begin{equation}
    p\!\left(x_{t+1:t+H}\mid x_{:t}\right),
\end{equation}
and to generate realistic continuations for long horizons $H$.

We work in a discrete latent space.
Let $\mathcal{E}_{\psi}$ and $\mathcal{D}_{\psi}$ denote a tokenizer encoder and decoder.
For an input segment $x$ we compute discrete codes and decode back to the signal domain:
\begin{equation}
    y = \mathcal{E}_{\psi}(x)\in\{0,\dots,K\!-\!1\}^{L},
    \qquad
    \hat{x} = \mathcal{D}_{\psi}(y),
\end{equation}
where $K$ is the codebook size and $L$ is the flattened token length.
We then train an autoregressive model $p_{\theta}(y)$ with next-token prediction and generate in token space before decoding.

\subsection{Tokenization: \texttt{BrainTokMix}}
A good MEG tokenizer must trade off three conflicting goals:
(i) high compression along \emph{time} and \emph{channels},
(ii) low reconstruction error,
and (iii) as few discrete symbols as possible (small vocabulary).
In our setting, the tokenizer defines the interface between a high-bandwidth continuous signal and a scalable Transformer, the better the tokenizer, the more ``language-like'' the downstream modeling becomes. A subtle but important point is that many vision-language models use continuous ``tokens'' (latent patches) only as conditioning for language generation.
In contrast, our downstream model is trained purely by next-token cross-entropy in the token space, which requires a discrete vocabulary.

\paragraph{Why not treat MEG as audio or video directly?}
MEG has native shape $T\times C$ (time $\times$ channels), unlike audio ($T$) or video ($T\times H\times W$).
Applying an audio codec independently to each channel would postpone cross-channel mixing until the Transformer and yields redundant tokens due to strong spatial correlations.
Rasterizing sensors into an image and treating MEG as a long low-resolution ``video'' is appealing because video tokenizers mix space and time jointly \citep{tang2024vidtok,cui2025emu35,agarwal2025cosmos}, but we found the resulting representation sparse and the tokenizer slower and less accurate for our setting.
These observations motivate a domain-specific tokenizer that (i) mixes channels early and (ii) compresses both time and spatial axes aggressively while remaining causal.

\paragraph{From BrainOmni to \texttt{BrainTokMix}.}
BrainOmni \citep{xiao2025brainomni} introduced a powerful sensor-aware neural tokenizer for EEG and MEG: it applies a SEANet-style codec \citep{defossez2022hifi} to each channel and then uses a dedicated sensor module to mix across sensors before quantization.
We adopt two core ingredients from this line of work: (i) a causal SEANet encoder-decoder backbone and (ii) their reconstruction and frequency-domain objectives (Eq.~\ref{eq:tokloss}).
In our MEG source-space regime, we can simplify the tokenizer and move mixing into the convolutional backbone.

\texttt{BrainTokMix} removes BrainOmni's sensor encoder and per-window sensor attention, sets sensor embeddings to zero, and performs spatiotemporal mixing via multichannel causal convolutions.
This yields an end-to-end causal codec that is easier to train efficiently (no batching over channels, no lstm over time, no attention over the $C$-sensor axis, and no metadata path) and produces discrete tokens that summarize joint spatiotemporal structure.

\subsubsection{Channel-mixing SEANet backbone}
Given a windowed multichannel recording $x \in \mathbb{R}^{C \times L_w}$, we use a strictly causal SEANet encoder-decoder \citep{defossez2022hifi}.
Concretely, the encoder consists of an initial causal convolution, two strided downsampling blocks (ratios $(2,2)$; overall hop length $r=4$), residual blocks with two residual layers.
This maps each window to a latent sequence $y \in \mathbb{R}^{T_w \times n_{\mathrm{dim}}}$ with $T_w=L_w/r$ and $n_{\mathrm{dim}}=4096$.

We then reshape the latent dimension into $n_{\mathrm{neuro}}$ streams:
$y_t \in \mathbb{R}^{n_{\mathrm{neuro}} \times d}$ where $n_{\mathrm{neuro}}=4$ and $d = n_{\mathrm{dim}}/n_{\mathrm{neuro}} = 1024$.
Intuitively, the model first performs spatiotemporal compression in a latent space, and the split exposes a small latent “spatial” axis that the downstream
Transformer can attend over. The decoder inverts this step by concatenating the streams back into a $4096$-D latent per timestep and applying a causal SEANet decoder to reconstruct $x$.

\paragraph{Residual vector quantization.}
Each latent vector $z_{h,t}\in\mathbb{R}^{d}$ (stream $h$, time $t$) is discretized using a $Q$-stage residual vector quantizer (RVQ) \citep{defossez2022hifi}.
Let $e^{(q)}_k$ denote the $k$-th code vector at RVQ level $q$.
RVQ selects indices sequentially on the residual: $r^{(q)} = r^{(q-1)} - e^{(q)}_{k^{(q)}}$, with $r^{(0)} = z_{h,t}$. The decoder receives the quantized latents $\tilde{z}$ (sum of $e^{(q)}_{k^{(q)}}$ across levels) and reconstructs the input.

\subsubsection{Tokenizer training}
The tokenizer is trained end-to-end to reconstruct the input while encouraging informative discrete codes.
Given reconstruction $\hat{x}$, we minimize the same loss designed for the BrainOmni tokenizer:
\begin{equation}
\label{eq:tokloss}
    \mathcal{L}
    = \lVert x-\hat{x}\rVert_1
    + \exp\!\left(-\mathrm{pcc}(x,\hat{x})\right)
    + \mathcal{L}_{\text{com}}
    + \mathcal{L}_{\text{amp}}
    + \tfrac{1}{2}\mathcal{L}_{\text{phi}},
\end{equation}
where $\mathrm{pcc}$ is the (channel-averaged) Pearson correlation coefficient, $\mathcal{L}_{\text{com}}$ is the RVQ commitment penalty, and $\mathcal{L}_{\text{amp}}$/$\mathcal{L}_{\text{phi}}$ compute the L1 loss between input and reconstruction FFT magnitudes and phases, respectively.
We train the tokenizer on 10.24\,s windows and then freeze it for autoregressive Transformer training.

For a segment of length $T$ (divisible by $L_w$), tokenization produces RVQ indices
\begin{equation}
    c_{t,h,q}\in\{0,\dots,K-1\},
\end{equation}
where $t\in\{1,\dots,T'\}$, $h\in\{1,\dots,n_{\text{neuro}}\}$, and $q\in\{1,\dots,Q\}$.  $T'=T/r$ is the downsampled time length.
The flattened token length is $L=T' n_{\text{neuro}} Q$, corresponding to a token rate
\begin{equation}
    \text{tokens/s} = f_s \cdot \frac{n_{\text{neuro}}Q}{r} = 100\cdot\frac{4\cdot 4}{4} = 400.
\end{equation}
This compression is what makes minute-scale contexts feasible for Transformers. Compared to flattening amplitude-quantized tokens at the full temporal and spatial dimensions this achieves a 17x compression ratio, and it is only 4x higher compared to folding the full spatial dimension into the batch or embedding, i.e. having $f_s=100$ tokens/s.

\subsection{Autoregressive modeling: \texttt{FlatGPT}}

A practical modeling question is how to represent a spatiotemporal signal in a decoder-only Transformer, which expects inputs shaped as $(\text{batch}, \text{length}, \text{embedding})$.
There are three natural options:
(1) put channels in the batch (yielding channel-independent models),
(2) put channels in the embedding (forcing the model to predict all spatial tokens for a time step jointly, without attention over them),
or (3) serialize/flatten spatiotemporal axes into the sequence.
We follow option (3), consistent with token-stream video models \citep{cui2025emu35,agarwal2025cosmos}: flattening permits full attention across both time and latent spatial streams under a standard causal mask.
While this imposes an arbitrary order over the latent spatial axis, (i) the axis is small ($n_{\text{neuro}}=4$), and (ii) we preserve its identity through axis-aware positional encodings (\cref{sec:mrope}).

\subsubsection{Flattening and next-token training}
We serialize the token grid by iterating RVQ level $q$ fastest:
\begin{align}
    i = \bigl((t-1)n_{\text{neuro}} + (h-1)\bigr)Q + q,
\end{align}
with $y_i \equiv c_{t,h,q}$ and $ L=T'n_{\text{neuro}}Q$. We then train a causal Transformer to model
\begin{equation}
    p_{\theta}(y) = \prod_{i=1}^{L} p_{\theta}\!\left(y_i \mid y_{<i}\right)
\end{equation}
with the standard teacher-forced cross-entropy loss
\begin{equation}
\label{eq:celoss}
    \mathcal{L}_{\text{AR}}(\theta)
    = -\sum_{i=1}^{L-1} \log p_{\theta}\!\left(y_{i+1}\mid y_{\le i}\right).
\end{equation}

\subsubsection{Transformer backbone and MRoPE}
\label{sec:mrope}
Once MEG is represented as a 3D token grid $(T',H',W')=(T/r,\;n_{\text{neuro}},\;Q)$, we can reuse video-capable Transformers.
We instantiate a Qwen-2.5-VL-style text Transformer \citep{qwen2024qwen25,bai2025qwen25vl} because it supports multimodal rotary position embeddings (MRoPE) used for flattened video tokens.

For each serialized token corresponding to $(t,h,q)$ we provide a 3-tuple position id
\begin{equation}
    \mathbf{p}_i = (p_i^{(t)}, p_i^{(h)}, p_i^{(w)}) = (t,\;h,\;q),
\end{equation}
stacked as $\mathbf{p}\in\mathbb{N}^{3\times B\times L}$.
MRoPE applies rotary embeddings to axis-specific subspaces of each attention head, allowing the model to reason about time, space, and even residual code levels distinctly while still using full attention.

\subsubsection{\texttt{FlatGPT} with RVQ-aware embeddings}
Our \texttt{FlatGPT} implementation is a thin wrapper that composes an arbitrary tokenizer with an arbitrary HuggingFace\footnote{https://huggingface.co/docs/transformers/en/index} decoder-only Transformer.
In our main configuration, we handle RVQ levels explicitly:
we use $Q$ separate embedding tables $\{E^{(q)}\}_{q=1}^Q$ so that the token embedding depends on the RVQ level,
\begin{equation}
    \mathrm{emb}(y_i) = E^{(q)}_{y_i} \in \mathbb{R}^{d_{\text{model}}}
\end{equation}
The output head is tied to the embedding weights with a one-step cyclic shift across RVQ levels, i.e. the head processing input from RVQ level 0 is tied to the embedding of RVQ level 1. This matches the fixed ordering of RVQ indices within each $(t,h)$ group and keeps parameters minimal. Total vocab size is $Q\times K$. This per-level vocabulary was crucial for good generation.

\subsubsection{Generation with a sliding KV cache}
To generate long continuations, we encode the provided context into tokens, autoregressively sample future tokens from $p_{\theta}$, and decode the generated tokens with $\mathcal{D}_{\psi}$.
Because rollouts can exceed the model's nominal context length, we use KV-cached decoding with a sliding-window approach: at generation time we keep a maximum of N context tokens (varies by experiment), and once this is reached we slide at the rate of the tokenizer encoding window (4096 tokens, 10.24\,s), refill the KV cache, then generate with caching up to N again, supporting multi-minute conditional generation efficiently. We align the stride to full tokenizer windows to avoid shifting RoPE position embeddings for partial windows; window boundaries can still induce subtle boundary effects, but we found this does not affect generation quality.

We note that \texttt{FlatGPT} is compatible with
context-scaling techniques (sparse attention, state-space models, RoPE curriculum), since the method’s core is simply “MEG $\rightarrow$ tokens $\rightarrow$ causal decoder”.

\subsection{Evaluation of long-horizon rollouts}

\paragraph{Rollout protocol.}
We sample full held-out sessions and form a context of 61.44\,s (start of session), followed by a continuation, with total evaluation segments of 296.96\,s (4.95\,min). Since many resting-state sessions are 5-minutes long this ensures we can include all sessions in our evaluations.
We generate one rollout per context. In a single analysis contexts are always drawn from a single task type from MOUS, i.e. rest, visual, or auditory, which makes our swapped controls rigorous.

\paragraph{Sliding-window on-manifold stability.}
We compute feature summaries on 30\,s windows with 5\,s stride for both generated and real continuations. In our main analyses we show: $1/f$ exponent, channel-covariance eigenvalue entropy, Welch PSD centroid, and an $\alpha$-bandpower ratio; with additional band-specific spectral metrics, long-range autocorrelation statistics (DFA/Hurst), and cross-channel connectivity summaries (covariance/coherence) in the Appendix.
To obtain an interpretable scalar curve we report an \emph{out-of-envelope rate} (OER): for each metric and window, we compute the 5--95\% envelope of real continuations and measure the fraction of generated runs outside it.

\paragraph{Conditional specificity.}

Let a test segment $i$ be split into a context $c_i\in\mathbb{R}^{C\times T_c}$ and its ground-truth continuation $y_i\in\mathbb{R}^{C\times T_y}$.
Conditioned on $c_i$, the model samples an open-loop continuation $x_i \sim p_\theta(\,\cdot\,|c_i)$.
For a prefix time $\tau\in(0,T_y]$ we write $x_i^{\le \tau}$ and $y_i^{\le \tau}$ for the first $\tau$ seconds of the \emph{continuation} (excluding the context).
We embed each prefix into a feature space $\phi(\cdot)$ (e.g., spectral and long-range statistics) and evaluate a distance $d(\cdot,\cdot)$, producing a prefix-divergence curve.

To disentangle conditional specificity from unconditional realism, we pair each context $i$ with a task-matched partner index $j=\pi(i)\neq i$ (i.e. different test session) and define the following per-context controls:
\begin{align}
D^{\textsc{correct}}_i(\tau)
&= d\!\left(\phi\!\left(x_i^{\le \tau}\right),\, \phi\!\left(y_i^{\le \tau}\right)\right), \\
D^{\textsc{prompt-swap}}_i(\tau)
&= d\!\left(\phi\!\left(x_{j}^{\le \tau}\right),\, \phi\!\left(y_i^{\le \tau}\right)\right) \\
D^{\textsc{target-swap}}_i(\tau)
&= d\!\left(\phi\!\left(x_i^{\le \tau}\right),\, \phi\!\left(y_{j}^{\le \tau}\right)\right) \\
D^{\textsc{real-real}}_i(\tau)
&= d\!\left(\phi\!\left(y_{j}^{\le \tau}\right),\, \phi\!\left(y_i^{\le \tau}\right)\right)
\label{eq:prefix_divergence_controls}
\end{align}

This isolates whether generations are (i) closer to the correct continuation than swapped baselines and (ii) calibrated relative to intrinsic real-data variability (variability calibration), rather than being merely on-manifold.
We summarize paired effects at the context level using bootstrap confidence intervals (5000 resamples) and Wilcoxon signed-rank tests.

\section{Experimental Setup}
\label{sec:experiments}

\subsection{Datasets and preprocessing}
\label{sec:datasets}
We train and evaluate on three public MEG datasets that differ in acquisition hardware and protocol:
CamCAN \citep{taylor2017camcan},
OMEGA \citep{niso2016omega},
and MOUS \citep{schoffelen2019mous}.
All recordings are converted to a common representation of $C=68$ source-space regions (Desikan--Killiany parcels \citep{desikan2006automated}) sampled at $f_s=100$\,Hz, yielding a fixed channel set across datasets, enabling direct cross-dataset training and evaluation.

\paragraph{Stage 1: preprocessing and source projection.}
We use an OSL \citep{gohil2023osl}/MNE-Python \citep{gramfort2013meg} preprocessing pipeline.
For CamCAN we apply Maxwell filtering \citep{taulu2006spatiotemporal}, for MOUS and OMEGA we apply gradient compensation (grade 3).
Then we run a minimal pipeline for each dataset consisting of a causal notch filter at the line noise frequency, then a causal bandpass filter between 1 and 50\,Hz, and causal resampling to 100\,Hz.
Bad channel detection is run and, when metadata exist, bad channels are interpolated.
We project sensor data to the \texttt{fsaverage} template and extract ROI time courses, yielding a consistent 68-channel source-space signal per session (see Appendix~\ref{app:preproc} for details).
While this does not give the most accurate source localization, we did not have access to subject MRIs for each dataset; for the purpose of cross-dataset generative modeling, a consistent and conservative projection is preferable to dataset-specific pipelines.

\paragraph{Stage 2: session cleaning.}
We apply robust normalization per session and channel using scikit-learn's \texttt{RobustScaler} (median/IQR; defaults) \citep{pedregosa2011scikit}.
We split the signal into fixed windows, drop windows whose standard deviation exceeds a threshold, and discard sessions with too many bad windows.
In our runs we use 5\,s windows, a standard-deviation threshold of 1.5, and discard sessions with more than 20\% bad windows.
We clip remaining samples to $[-10,10]$ (in normalized units) and save contiguous ``good'' segments that are at least 60\,s long, discarding any shorter segments.

\paragraph{Train/validation/test splits.}
We train \emph{both} the tokenizer and Transformer on CamCAN+OMEGA and hold out MOUS entirely for validation and testing.
MOUS subjects are split 50/50 into val/test with a fixed random seed. After cleaning, this yields 2684 training sessions from CamCAN and 1719 from OMEGA (420 hours, 6$\times 10^8$ tokens), and 198/191 MOUS sessions for validation/testing (roughly 70 hours each).
See Appendix~\cref{tab:data_breakdown} for a summary table.

\subsection{\texttt{BrainTokMix} tokenizer setup and training}
\label{sec:tok_train}
\texttt{BrainTokMix} uses a causal SEANet with window length $L_w=1024$ samples (10.24\,s), downsampling ratios $(2,2)$, $n_{\mathrm{filters}}=1024$, $n_{\mathrm{dim}}=4096$, and $n_{\mathrm{neuro}}=4$ streams (token width $d=1024$).
For the RVQ we use $Q=4$ codebooks, codebook size $K=16384$, and code dimension $1024$. Full model size is 294M parameters.

We train the tokenizer with the objective in Eq.~\ref{eq:tokloss} using 10.24\,s examples.
We use AdamW \citep{loshchilov2017decoupled} (lr $5\times10^{-5}$, weight decay $10^{-2}$), linear warmup over 300 steps, and gradient clipping of 1.0.
The batch size is 480 windows, i.e., $480\times10.24s\approx82$ minutes of MEG per optimization step.
We train for 20 epochs, which takes about 5 hours on a B200 GPU. VQVAEs are hard to overfit even without regularization and we simply stop training when improvement over one epoch is marginal. In additional runs, increasing tokenizer capacity improved reconstruction, but we found the gains modest relative to the extra compute and therefore use this 294M configuration as a practical trade-off (see Appendix~\ref{app:discussion_extended}).
After training, we freeze the tokenizer weights for all autoregressive experiments.

\subsection{\texttt{FlatGPT} architecture and training}
\label{sec:model_arch}
We instantiate \texttt{FlatGPT} with a Qwen2.5-VL-style decoder-only Transformer backbone\footnote{https://huggingface.co/docs/transformers/en/index} and train it from scratch on \texttt{BrainTokMix} tokens.
The backbone has 12 layers, hidden size 1200, 10 attention heads (2 KV heads), head dimension 120, and MLP width 4560, in total 336M parameters.

We use AdamW with learning rate $2\times10^{-4}$, weight decay 0.1, linear warmup over 2000 steps, and gradient clipping at 1.0.
With a token rate of 400 tokens/s, a 61.44\,s example contains 24{,}576 tokens.
At batch size 8 this corresponds to 196{,}608 tokens per optimization step. We use early-stopping on the MOUS validation sessions, resulting in 8 epochs. 1 epoch takes about 1.7 hours on a B200 GPU. Both the tokenizer and backbone are trained with BF16 mixed precision and \texttt{torch.compile}. For the Qwen backbone we found the cuDNN sdpa backend the fastest\footnote{https://docs.pytorch.org/docs/stable/backends.html}.

\subsection{Generation and evaluation protocol}
\label{sec:eval_setup}
All results are reported on the MOUS \textbf{test} split.
For each evaluation run and task type we sample all available session segments and use the first 61.44\,s as context and a total length of 296.96\,s, i.e., 235.52\,s of open-loop continuation. We generate 1 rollout (94k tokens) per context with temperature $=1.0$ and top-$p=1.0$ for sampling, i.e., pure multinomial sampling; alternative sampling heuristics (e.g., lower top-$p$ or per-RVQ-level temperature schedules) generally worsened rollouts.

We evaluate MOUS task types independently: auditory/listening ($n=41$), visual/reading ($n=34$), and resting-state ($n=71$). The sum of these is less than the number of test sessions due to not all sessions having a contiguous 5-minute beginning after the session cleaning. Our setting is intentionally stringent: MOUS is fully held out from tokenizer and model training, stimulus annotations are not used, and the rollout horizon is $\approx3.8\times$ longer than the model and conditioning context, so uncertainty compounds under intrinsic stochasticity and noise.

We compute sliding-window summaries of generated vs real runs on 30\,s windows with 5\,s stride. We compute \emph{prefix divergence curves} at prefix times $\tau\in\{20,40,60,80,100,150,200,250\}\,$s, plus the max prefix. In our main analyses we use the following distances and features: 1. normalized L2 distance between channel-covariance matrices, 2. Jensen-Shannon divergence between PSD distributions, 3. normalized L2 distance between broadband coherence matrices, with many additional band-specific and auto-correlation/connectivity metrics in the Appendix (e.g., DFA/Hurst exponents, bandpower ratios, and spatial connectivity summaries).

\section{Results}
\label{sec:results}

\subsection{\texttt{BrainTokMix} reconstruction fidelity}

Because \texttt{FlatGPT} operates purely in \texttt{BrainTokMix} token space, tokenizer reconstruction bounds downstream signal fidelity.
On held-out MOUS, \texttt{BrainTokMix} achieves low reconstruction error ($\mathrm{MAE}=0.2$, $\mathrm{PCC}=0.944$) with high channel-wise correlation and near-uniform code usage (Appendix~\cref{tab:tok_metrics}). Note that this is much better than what was reported in \citet{xiao2025brainomni}, likely due to improved model expressivity and scaling of model size.
A small but consistent attenuation of high-frequency power is visible in the reconstructed PSD (Appendix \cref{fig:tok_fidelity}); this likely contributes to, slightly reduced gamma-band power in long-horizon generations. Pushing either temporal or spatial reduction further (i.e. 2$\times$ our current setup) resulted in worse reconstruction quality, and a maximum $\mathrm{PCC}$ of 0.9. Increasing the number of RVQ levels mitigates this, but then the actual tokens/s is not reduced (due to our flattening approach). Therefore our current setup is quite close to optimality in terms of the reduction--reconstruction trade-off.

\subsection{On-manifold stability over 4-minute rollouts}

\begin{figure}[!t]
\centering
\includegraphics[width=0.48\textwidth]{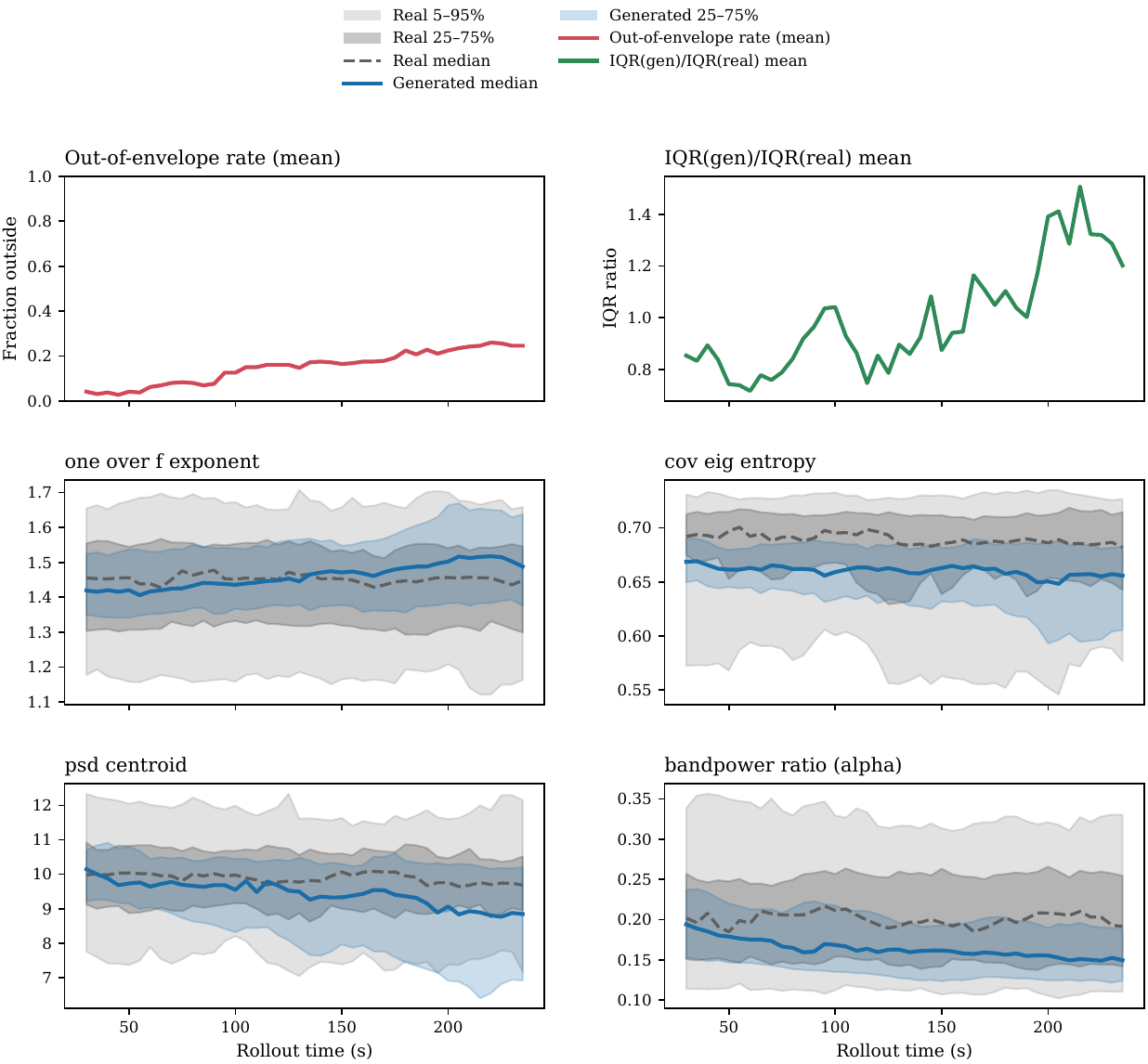}
\caption{\textbf{On-manifold stability for resting-state rollouts (MOUS test).} Gray bands show the real 5--95\% and 25--75\% envelopes; blue shows the generated distribution across contexts. Top 2: mean OER and IQR ratio. Bottom 4: feature stability.}
\label{fig:window_rest}
\end{figure}

We first test whether open-loop generation drifts off-manifold using the \emph{out-of-envelope rate} (OER; \cref{sec:eval_setup}).
Across all three tasks (rest shown in Figure~\ref{fig:window_rest}), generated windows largely remain within the distributional envelope of real windows for key neurophysiological summaries, with drift accumulating gradually over the 4\,min continuation. Full stability plots for each task (including band-specific spectral metrics, DFA/Hurst exponents capturing long-range autocorrelation) are provided in Appendix \cref{fig:window_full_aud,fig:window_full_vis,fig:window_full_rest}.

\subsection{Conditional specificity via prefix divergence}

We next test whether generations are \emph{conditionally specific} to the correct prompt and continuation using prefix divergence curves with task-matched swap controls (Eq.\,\ref{eq:prefix_divergence_controls}).
Figure~\ref{fig:divergence_rest} shows our main metrics for rest only, with all metrics and task types in Appendix \cref{fig:div_full_aud,fig:div_full_vis,fig:div_full_rest}. Correct
generations are consistently closer to the true continuation
than task-matched controls and the real-real
baseline, but the gap does decrease with increased rollout horizon.
Table~\ref{tab:prefix_main} quantifies these gaps at the end of the rollout, supporting both conditional specificity (prompt-swap) and variability calibration against natural real variability (real-real).

Prompt dependence persists far beyond the conditioning window: at 235.5\,s generated, correct continuations reduce covariance distance by 0.088--0.130 relative to prompt-swap controls, and remain closer than the real-real baseline (\cref{tab:prefix_main}; target-swap shown in Appendix~\cref{tab:targetswap_stats}).
Qualitatively, long-horizon generations preserve global structure: average covariance heatmaps and PSDs closely match ground-truth across tasks (Appendix ~\cref{fig:global_aud,fig:global_vis,fig:global_rest}).
Representative time-series and STFT rollouts qualitatively resemble their targets without obvious artifacts (Appendix \cref{fig:qual_rollouts,fig:qual_rollouts_rest}).

\begin{figure}[!t]
\centering
\includegraphics[width=0.48\textwidth]{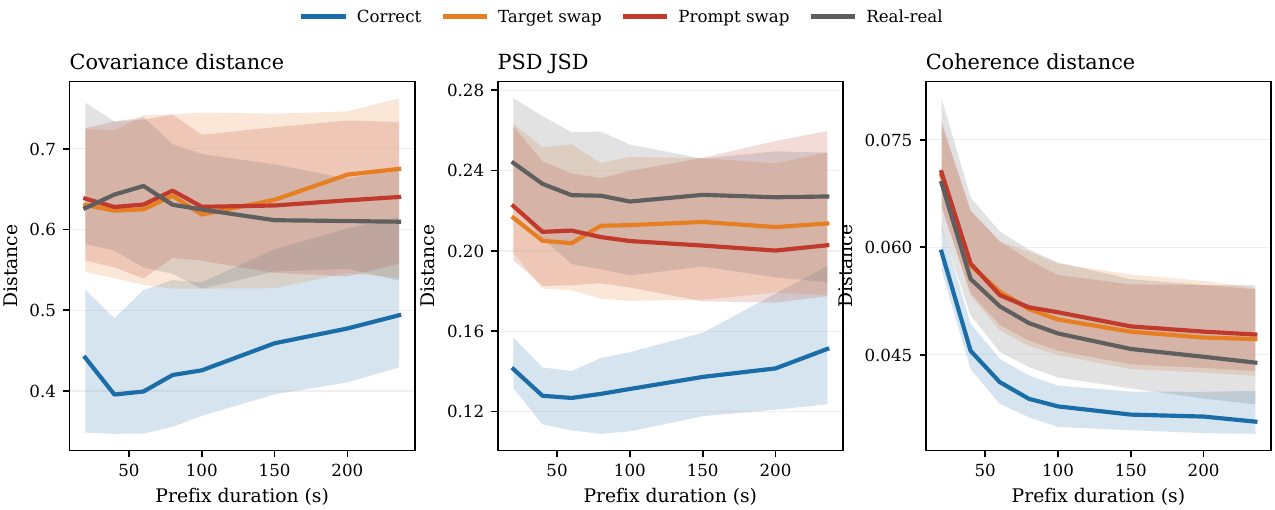}
\caption{\textbf{Conditional specificity for resting-state rollouts (MOUS test).} Prefix divergence over increasing generated duration $\tau$ for the correct pairing (blue) versus prompt-swap (red) target-swap (orange) controls and a real-real baseline (gray). Shaded regions show interquartile ranges across contexts.}
\label{fig:divergence_rest}
\end{figure}

\begin{table*}[t]
\centering
\caption{\textbf{Conditional specificity at 235.5\,s.} We report paired median improvements $\Delta$ (control $-$ correct) with 95\% bootstrap CIs on the MOUS test set. \emph{Prompt-swap} tests dependence on the correct conditioning prefix. \emph{Real-real} compares to a task-matched baseline distance between two real continuations (variability calibration). Larger $\Delta$ means the correct generation is closer to the target than the control.}
\label{tab:prefix_main}
\small
\setlength{\tabcolsep}{5pt}
\begin{tabular}{l c l c c c}
\toprule
Task & $n$ & Control & Covariance distance & PSD JSD & Coherence distance \\
\midrule
Auditory & 41 & Prompt-swap & 0.130 [0.075,0.235] & 0.048 [0.024,0.073] & 0.0065 [0.0052,0.0086] \\
 &  & Real-real & 0.097 [0.062,0.181] & 0.047 [0.025,0.054] & 0.0059 [0.0030,0.0077] \\
 \addlinespace
Visual & 34 & Prompt-swap & 0.096 [0.027,0.129] & 0.027 [0.011,0.052] & 0.0099 [0.0083,0.0112] \\
 &  & Real-real & 0.076 [0.001,0.149] & 0.046 [0.024,0.067] & 0.0065 [0.0055,0.0122] \\
 \addlinespace
Rest & 71 & Prompt-swap & 0.088 [0.063,0.173] & 0.062 [0.042,0.077] & 0.0096 [0.0080,0.0115] \\
 &  & Real-real & 0.098 [0.046,0.135] & 0.074 [0.050,0.080] & 0.0074 [0.0053,0.0096] \\
\bottomrule
\end{tabular}
\end{table*}

\subsection{Context length ablation}

Shorter model and conditioning contexts degrade both stability and conditional specificity.
Reducing the context from 61.44\,s to 30.72\,s increases mean OER on nearly all stability metrics, and shrinks the correct-vs-swap gaps (Appendix \cref{sec:ablation_30s}).
\cref{tab:ablation_stats} quantifies how prompt-swap separation at 235.5\,s weakens with a 30\,s context; notably, on the visual task the PSD-JSD gap is no longer statistically significant. Teacher-forced loss also decreases slightly with context length (Appendix~\cref{fig:token_summary_aud}).

\begin{table}[t]
\centering
\caption{\textbf{Prompt-swap separation at 235.5\,s: 60\,s vs. 30\,s context.} We report paired median improvements $\Delta$ (control $-$ correct) for the prompt-swap control and the corresponding $p$-value for the 30\,s context (paired Wilcoxon).}
\label{tab:ablation_stats}
\small
\setlength{\tabcolsep}{6pt}
\begin{tabular}{l l c c c}
\toprule
Task & Metric & $\Delta$ (60\,s) & $\Delta$ (30\,s) & $p$ (30\,s) \\
\midrule
Auditory & covariance & 0.130 & 0.087 & 5.4e-3 \\
Auditory & PSD-JSD & 0.048 & 0.045 & 7.9e-4 \\
Auditory & coherence & 0.0065 & 0.0048 & 1.9e-10 \\
Visual & covariance & 0.096 & 0.050 & 0.041 \\
Visual & PSD-JSD & 0.027 & 0.009 & 0.37 \\
Visual & coherence & 0.0099 & 0.0052 & 1.0e-8 \\
Rest & covariance & 0.088 & 0.086 & 9.0e-6 \\
Rest & PSD-JSD & 0.062 & 0.044 & 1.3e-6 \\
Rest & coherence & 0.0096 & 0.0062 & 3.6e-13 \\
\bottomrule
\end{tabular}
\end{table}

\section{Discussion}

We introduced \texttt{BrainTokMix}, a causal spatiotemporal RVQ tokenizer for fixed-channel-order MEG, and \texttt{FlatGPT}, a decoder-only Transformer trained on the resulting flattened token stream.
Training the tokenizer \emph{and} Transformer backbone on CamCAN+OMEGA and evaluating solely on held-out MOUS, \texttt{FlatGPT} can condition on 1 minute of context and generate at least 4 minutes of open-loop continuation while (i) largely staying within the real-data envelope of neurophysiological summaries, and (ii) remaining measurably dependent on the specific ``prompt".

\paragraph{Why tokenization matters.}
The tokenizer sets sequence length and determines whether the downstream autoregressive distribution is learnable.
In our experiments, BrainOmni-style tokenization achieved comparable reconstruction quality (to \texttt{BrainTokMix}) at the same reductions but was roughly $\sim3\times$ slower to train, and a VidTok \citep{tang2024vidtok} baseline was substantially slower and reached only 0.90 PCC.
In additional experiments (not shown), we found diminishing returns from simply increasing codebook size, whereas adding RVQ levels can improve fidelity but seems to make later levels harder to predict and destabilize long rollouts. Transformer context scaling also had diminishing returns: a long-context curriculum (progressively increasing context length/RoPE parameters up to 160\,s) did not improve rollout metrics. More practical takeaways and lessons learned are discussed in Appendix~\ref{app:discussion_extended}

\paragraph{Limitations.}
Our evaluation uses task-matched controls under a deliberately hard and general setting (OOD MOUS, no stimulus annotations, 94k-token rollouts). However, we did not test stimulus-locked correctness, which is left for future work. While predicting the beginning of an evoked response without stimulus information is informationally underspecified, one interesting analysis (with our no-stimulus-label paradigm) would be to test evoked generation fidelity when supplying the model a short initial window after stimulus presentation.

Due to lack of good baselines we omitted baseline sweeps. To our knowledge, the recent MEG-GPT \citep{huang2025meggpt} is the only prior multi-channel brain foundation model demonstrated for open-loop generation, but it reports training and generation with an 800\,ms context and, in our setup, took roughly $10\times$ longer to train than \texttt{FlatGPT}; extending it to minute-scale contexts would be computationally prohibitive, so an apples-to-apples comparison is currently impractical. Classical AR/VAR baselines can match coarse PSD statistics, but fail to reproduce cross-channel covariance and transient events (see our time-series/STFT plots), making them weak comparators for conditional long-horizon generation \citep{csaky2024foundational}.

Better tokenizers that improve high-frequency fidelity, systematic scaling studies (data, model, context size), and multimodal conditioning on tokenized stimuli are promising next steps. Generative brain priors may also serve as a privileged latent for distillation or alignment.

\section*{Acknowledgements}
This research was fully funded by an AI Safety Grant from the Foresight Institute.

\section*{Impact Statement}

This work enables promptable, multi-minute neural signal generation that generalizes across datasets, opening new avenues for simulation, evaluation, and multimodal brain-stimulus modeling. Because neural data can be identifying and sensitive, and synthetic signals could be misused or overinterpreted, any deployment or release should follow strict consent safeguards and present distributional samples.

\bibliography{ml}
\bibliographystyle{icml2026}

\newpage
\appendix
\twocolumn

\section{Appendix}

\subsection{Extended Related Work}
\label{app:related_work}

\paragraph{Generative modeling and forecasting of electrophysiology.}
Beyond representation learning, there is growing interest in generative models that can synthesize realistic neural signals. EEGTrans uses a quantized autoencoder together with an autoregressive Transformer decoder to generate discrete EEG code sequences for data synthesis \citep{lim2024eegtrans}. MEG-GPT trains an autoregressive Transformer with next-step prediction on tokenized MEG region time courses, showing that generated signals match spatio-spectral properties and can improve downstream decoding \citep{huang2025meggpt}. In parallel, diffusion models and other continuous generative approaches have been explored for time-series generation and forecasting. Compared to these efforts, \texttt{FlatGPT} emphasizes (i) purely next-token objective over discrete MEG tokens through an efficient and scalable paradigm (ii) long-context conditioning through prompting rather than task labels, and (iii) stress-testing open-loop generations for stability and context specificity across datasets.

\paragraph{Time-series foundation models.}
Outside neuroscience, recent work has started to build general-purpose \emph{time-series foundation models} (TSFMs) by pretraining large Transformers on large corpora of heterogeneous time series and evaluating them in zero-/few-shot forecasting settings. Representative examples include decoder-only pretrained forecasters such as TimesFM \citep{tsfm_das2024timesfm} and TimeGPT \citep{tsfm_garza2023timegpt}, probabilistic TSFMs such as Lag-Llama \citep{tsfm_rasul2023lagllama}, and approaches that explicitly discretize values and apply language-model training, such as Chronos \citep{ansari2024chronos}. While the primary goal of TSFMs is typically accurate and transferable forecasting for generic (often low-dimensional) time series, \texttt{FlatGPT} targets a different axis: building a generative prior over high-bandwidth multichannel MEG and stress-testing \emph{open-loop} rollouts for long-horizon stability and prompt dependence.

\paragraph{Brain-language alignment and multimodal neural models.}
Generative models of neural signals are also motivated by downstream decoding tasks, such as reconstructing stimuli or behavior. For example, MEG can be used to decode continuous speech from non-invasive recordings \citep{defossez2022decoding}. More recently, several works treat brain activity as a ``foreign language'' by learning neural tokenizers and coupling them to LLM backbones. NeuroLM learns a text-aligned EEG tokenizer and uses instruction tuning for multi-task EEG inference \citep{jiang2024neurolm}. NeuGPT and fMRI-LM similarly aim to jointly model neural tokens and text to enable language-conditioned understanding from neural recordings \citep{yang2024neugpt,wei2025fmrilm}. Orthogonally, work in cognitive NLP studies representational alignment between LLMs and neural responses, including the effect of instruction tuning \citep{aw2023instruction} and evidence that model scaling and training choices can systematically increase alignment \citep{gao2025increasingalignment}. Related ``brain-tuning'' approaches fine-tune speech/language models directly on fMRI to induce brain-relevant semantics \citep{moussa2024braintuning}. These approaches typically rely on curated neural-text alignment or task supervision; \texttt{FlatGPT} is complementary in targeting an unsupervised generative prior over MEG dynamics, which could serve as a backbone for future multimodal conditioning or decoding.

\paragraph{Evaluating generative neural models.}
Unlike text or images, the realism of generated MEG cannot be judged visually, and models can match simple marginal statistics while failing to respect the conditioning prompt or long-range dynamics. Most prior work reports token reconstruction, masked prediction accuracy, or downstream decoding performance \citep{wang2024eegpt,jiang2024labram,xiao2025brainomni,huang2025meggpt}. To evaluate open-loop generation, we introduce metrics and controls that probe (i) distributional drift over long rollouts and (ii) context specificity via prompt swapping and permutation-style controls. This evaluation perspective mirrors how generative models are stress-tested in other modalities, but is adapted to the unique challenges of electrophysiology.

\subsection{Positioning relative to prior work}
\label{app:positioning}

\begin{table*}[!t]
\vspace{-0.25em}
\centering
\caption{High-level positioning of \texttt{FlatGPT} relative to closely related work.
Entries for prior work summarize the primary setting emphasized in each paper. \citep{huang2025meggpt,xiao2025brainomni,jiang2024labram,vetter2024generating}}
\label{tab:positioning}
\small
\begin{tabular}{lccccc}
\toprule
 & \textbf{\texttt{FlatGPT} (ours)} & \textbf{MEG-GPT}  & \textbf{BrainOmni}  & \textbf{LaBraM}  & \textbf{NTD}  \\
\midrule
Space & source & source & sensors & sensors & sensors \\
Tokenization & RVQ & lossless per-sample & RVQ & VQ & none \\
Training objective & autoregressive & autoregressive & masked recon & masked recon & denoising recon \\
Context modeled & 60s & 0.32s & 30s & 4-8s & 1-4s \\
Open-loop generation & 240s & 60s & N/A & N/A & 1-4s \\
Data & multi-dataset & single-dataset & multi-modality & multi-dataset & single-dataset \\
Generalization & held-out dataset & within-subject & held-out dataset & held-out dataset & within-subject \\
\bottomrule
\end{tabular}
\vspace{-0.5em}
\end{table*}

\subsection{Preprocessing details}
\label{app:preproc}

\begin{table}[!h]
\centering
\caption{\textbf{Cleaned dataset breakdown.} Session counts are after cleaning; hours/tokens are totals per split.}
\label{tab:data_breakdown}
\small
\setlength{\tabcolsep}{6pt}
\begin{tabular}{l l r r r}
\toprule
Dataset & Split & Sessions & Hours & Tokens \\
\midrule
CamCAN & train & 2684 & \multirow{2}{*}{420} & \multirow{2}{*}{$6.05\times 10^8$} \\
OMEGA  & train & 1719 &  &  \\
\midrule
MOUS & val  & 198 & 70 & $1.01\times 10^8$ \\
MOUS & test & 191 & 70 & $1.01\times 10^8$ \\
\bottomrule
\end{tabular}
\end{table}

We perform MRI-less coregistration to the \texttt{fsaverage} template using digitized fiducials and head-shape points (conservative ICP; MNE defaults) \citep{besl1992method,gramfort2013meg}.
We then compute an \texttt{ico5} forward model (BEM; \texttt{mindist}=3\,mm) and obtain dSPM minimum-norm source estimates (\texttt{snr}=3, \texttt{loose}=0.2, \texttt{depth}=0.8, ad-hoc noise covariance) with fixed normal orientation \citep{hamalainen1994interpreting,dale2000dynamic}.
Finally, we extract Desikan--Killiany ROI time courses (mode=\texttt{mean}; MNE default) and linearly detrend each ROI, yielding a consistent 68-channel source-space signal per session.

Table~\ref{tab:data_breakdown} summarizes the cleaned dataset sizes used in our experiments. Hours refer to the total duration of contiguous ``good'' segments retained after Stage~1--2 preprocessing (\cref{sec:datasets}). Token counts are obtained by multiplying hours by the tokenizer rate (400 tokens/s).

\subsection{Tokenizer diagnostics}
\begin{table}[!h]
\centering
\caption{\textbf{\texttt{BrainTokMix} reconstruction metrics on held-out MOUS}. Maximum codebook usage perplexity is 16{,}384.}
\label{tab:tok_metrics}
\small
\setlength{\tabcolsep}{6pt}
\begin{tabular}{l c}
\toprule
Metric & Value \\
\midrule
MAE $\downarrow$ & 0.203 \\
PCC $\uparrow$ & 0.944 \\
FFT amplitude error $\downarrow$ & 0.0835 \\
FFT angle error $\downarrow$ & 0.806 \\
Commit loss $\downarrow$ & $2.67\times10^{-4}$ \\
Codebook perplexity $\uparrow$ & 15,518 \\
\bottomrule
\end{tabular}
\end{table}

\begin{figure*}[!h]
\centering
\begin{subfigure}{0.49\linewidth}
\centering
\includegraphics[width=\linewidth]{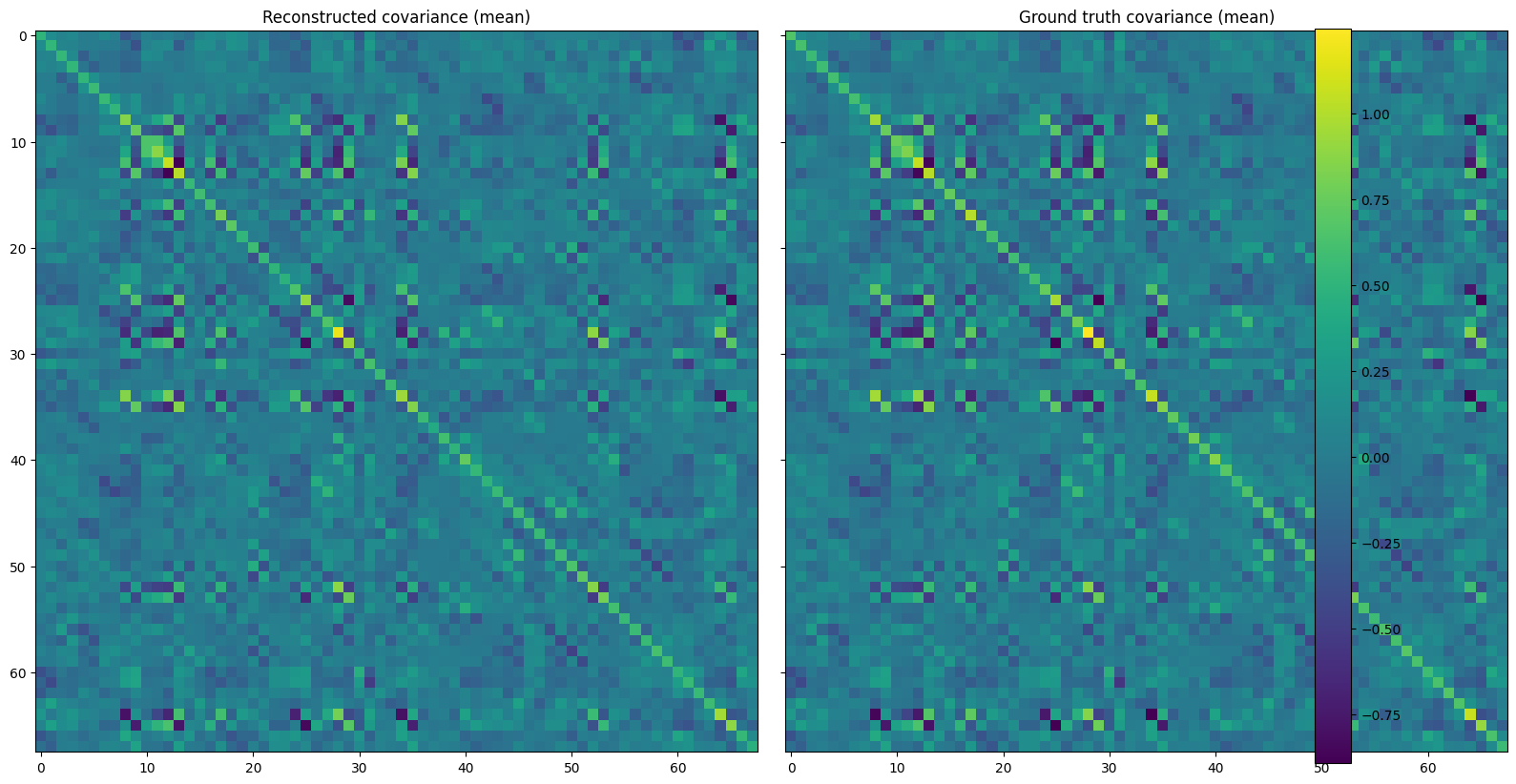}
\caption{Covariance structure averaged over held-out windows.}
\end{subfigure}
\hfill
\begin{subfigure}{0.49\linewidth}
\centering
\includegraphics[width=\linewidth]{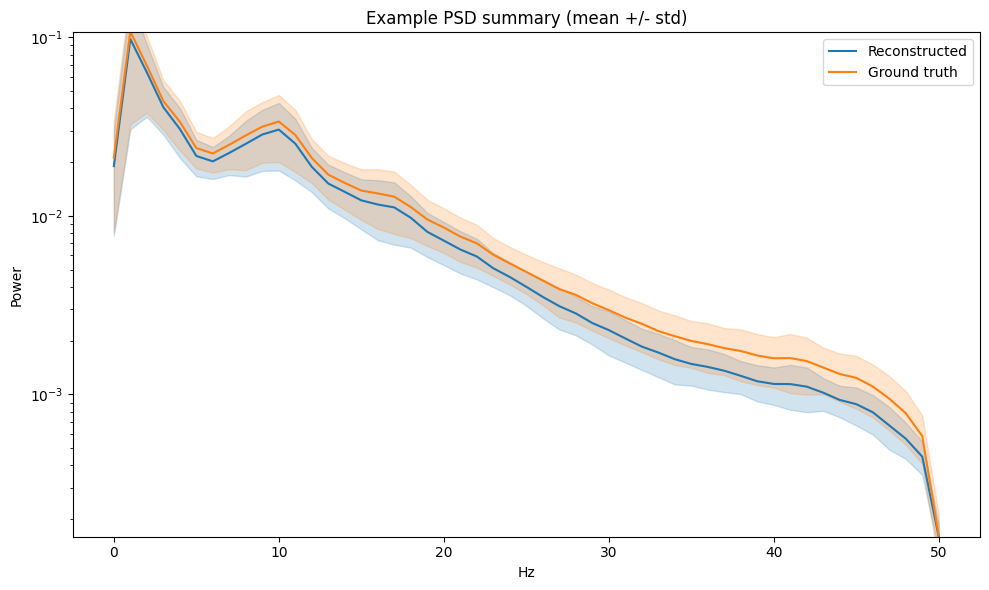}
\caption{Power spectra averaged over held-out windows and channels.}
\end{subfigure}
\caption{\textbf{\texttt{BrainTokMix} reconstruction preserves spatial and spectral statistics.} Reconstructions closely match target covariance and PSD across held-out MOUS windows, with mild attenuation at higher frequencies (likely contributing to slightly reduced gamma-band power downstream).}
\label{fig:tok_fidelity}
\end{figure*}

\subsection{Target-swap statistics}

\begin{table*}[!t]
\centering
\caption{\textbf{Target-swap control at 235.5\,s generated} (paired median $\Delta$ with 95\% bootstrap CI; MOUS test). $\Delta$ is reported as (target-swap $-$ correct), so larger is better.}
\label{tab:targetswap_stats}
\small
\setlength{\tabcolsep}{6pt}
\begin{tabular}{l c c c}
\toprule
Task & Covariance distance & PSD-JSD & Coherence distance \\
\midrule
Auditory & 0.119 [0.057,0.181] & 0.042 [0.031,0.056] & 0.0096 [0.0065,0.0118] \\
Visual   & 0.081 [0.058,0.115] & 0.037 [0.021,0.050] & 0.0085 [0.0042,0.0122] \\
Rest     & 0.135 [0.062,0.185] & 0.056 [0.049,0.074] & 0.0101 [0.0077,0.0116] \\
\bottomrule
\end{tabular}
\end{table*}

\subsection{Token-level loss vs. context length}

\begin{figure}[!h]
\centering
\begin{subfigure}{0.48\textwidth}
\centering
\includegraphics[width=\linewidth]{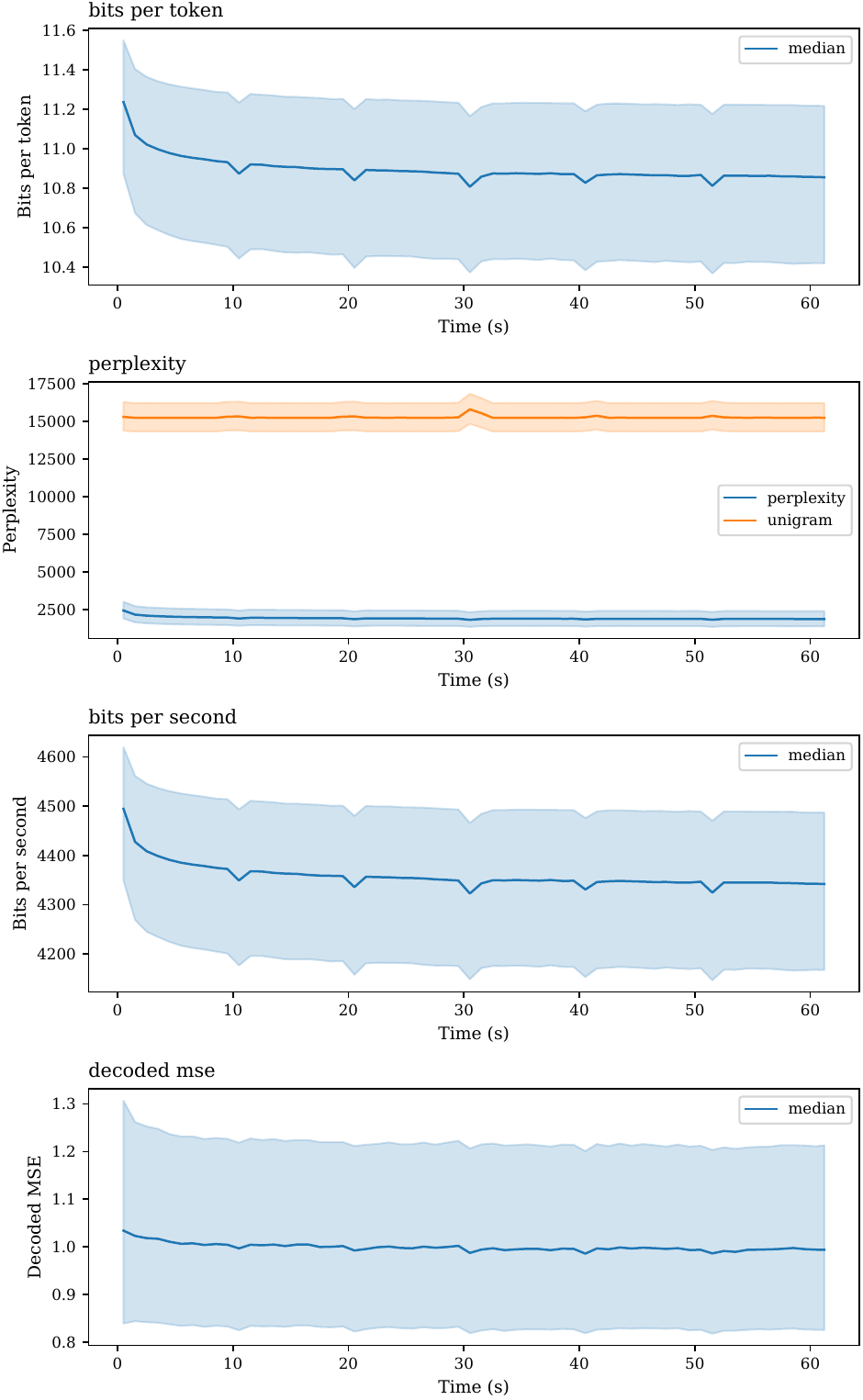}
\end{subfigure}
\caption{\textbf{Token-level prediction vs. available context on test data.}}
\label{fig:token_summary_aud}
\end{figure}

These teacher-forced summaries (Figure~\ref{fig:token_summary_aud}) quantify how next-token prediction improves as more real context is available.
The periodic ``sawtooth'' structure in bits-per-token and perplexity is due to the tokenizer window length (10.24\,s), which induces window-aligned shifts in token statistics.

\subsection{Extended Discussion}
\label{app:discussion_extended}

\paragraph{Tokenizer lessons: reconstruction, compression, and predictability must be balanced.}

We observed that window length is a real modeling constraint: longer windows help reconstruction by allowing the codec to use past context within the window, but windowing can introduce periodic effects in token statistics (reflected in the sawtooth token-loss curves; Appendix \cref{fig:token_summary_aud}).
Importantly, in OMEGA-only trials we obtained similar downstream generation results with a much shorter tokenizer window (1.28\,s), suggesting the model cannot ``cheat'' by relying on windowing; if anything, boundary effects make the autoregressive task harder.
While overlap-add decoding can reduce boundary artifacts for reconstruction, it is not available for open-loop autoregressive generation because overlapping regions would require future tokens.

We also tried several other modifications of BrainOmni, including interleaving temporal and spatial reductions, but training proved difficult. Compared to the original BrainOmni setup, removing channel-masking, denoising, and normalization improved reconstruction quality substantially. Since the tokenizer is not able to overfit anyway (due to the RVQ bottleneck and large reductions) we believe there is no need for these further regularization techniques.

\paragraph{Practical scaling notes for long-context MEG.}
A recurring theme in this work is that ``LLM-style simplicity'' is a feature: \texttt{FlatGPT} uses the standard next-token objective, standard decoder-only training, and standard KV-cached sampling with a sliding context window.
In our experience, scaling is most constrained by data heterogeneity rather than architectural novelty: getting a single model to train stably across hundreds of hours (420 after cleaning) and thousands of sessions spanning multiple scanners and tasks is challenging.

Sliding-window attention masks provided only modest training speedups and slightly degraded generation quality; we suspect this is because the flattened token stream interleaves channels/streams and benefits from full causal coupling to maintain covariance structure.
Transformer scale exhibited the expected compute trade-off: smaller backbones could reach comparable performance, but typically required more epochs to do so, reducing the effective compute savings.
Backbone choice also mattered: in our OMEGA-only trials, Qwen2.5 training was more stable than some alternative bases (including a Qwen3 variant, which produced noisier rollouts).

We have tried \texttt{FlatGPT} variants where the RVQ levels are folded (concatenating embeddings) into the hidden dimension of the Transformer to reduce sequence length and be predicted jointly at each step. While this does improve training speed substantially long rollouts were less stable, leading to early degeneration.

\subsection{Global metrics for 60\,s-context rollouts}

\begin{figure*}[!h]
\centering
\begin{subfigure}{0.49\linewidth}
\centering
\includegraphics[width=\linewidth]{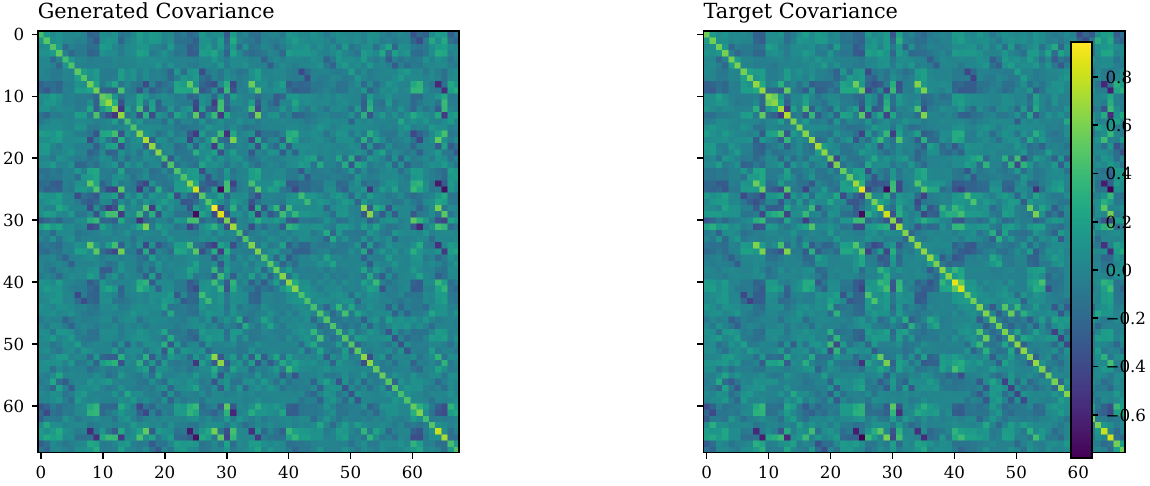}
\caption{Auditory: covariance.}
\end{subfigure}
\hfill
\begin{subfigure}{0.49\linewidth}
\centering
\includegraphics[width=\linewidth]{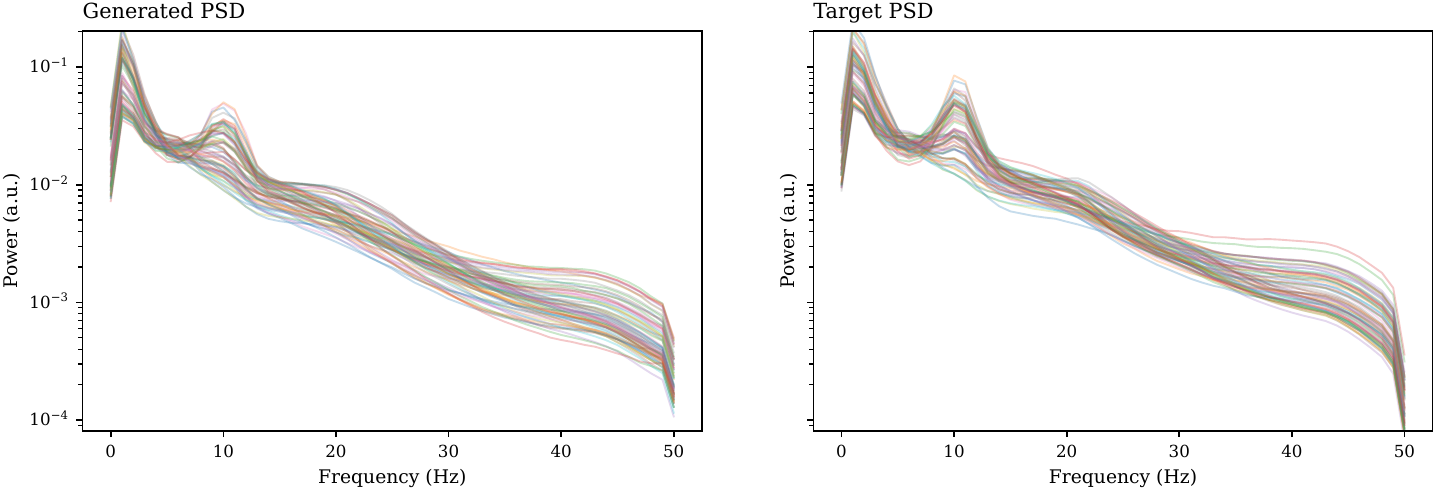}
\caption{Auditory: PSD.}
\end{subfigure}
\caption{\textbf{Global covariance and PSD for auditory rollouts (60\,s context).} Left: covariance heatmaps averaged over generated and target continuations. Right: channel PSDs (0--50\,Hz).}
\label{fig:global_aud}
\end{figure*}

\begin{figure*}[!h]
\centering
\begin{subfigure}{0.49\linewidth}
\centering
\includegraphics[width=\linewidth]{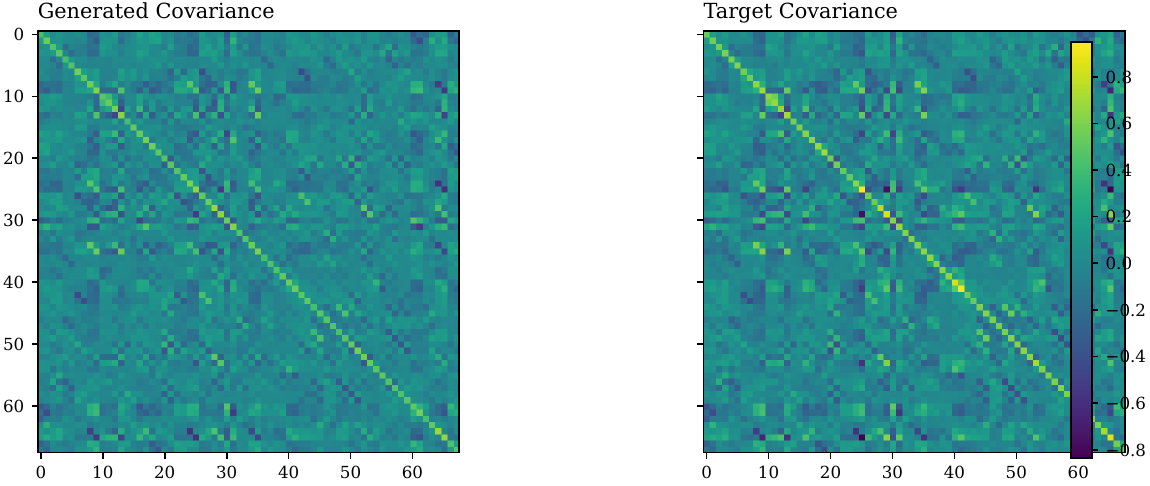}
\caption{Visual: covariance.}
\end{subfigure}
\hfill
\begin{subfigure}{0.49\linewidth}
\centering
\includegraphics[width=\linewidth]{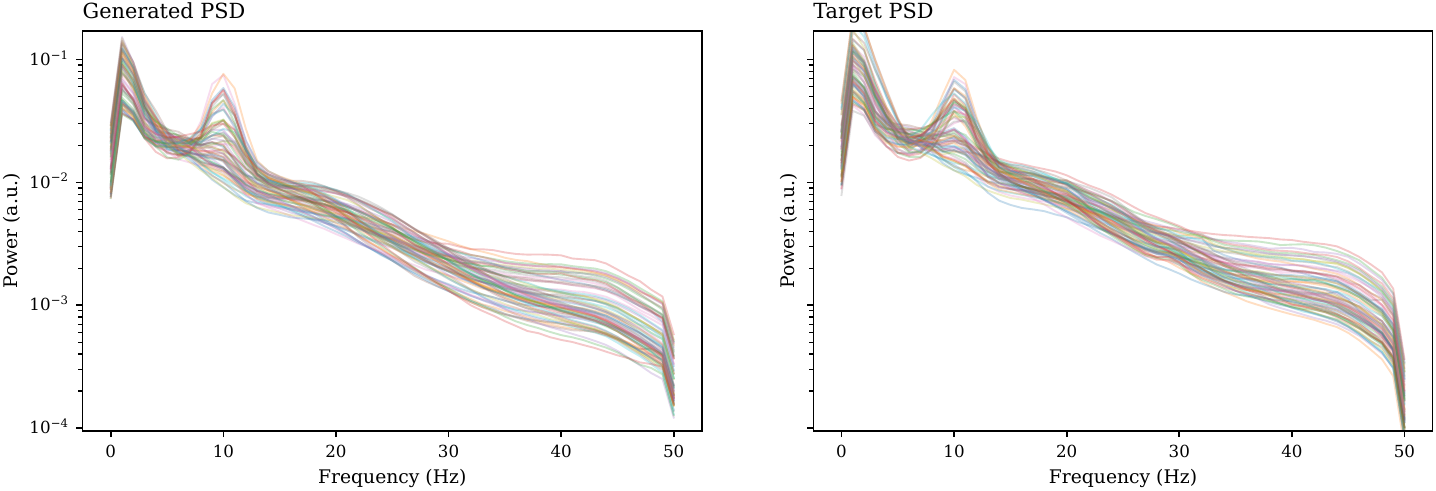}
\caption{Visual: PSD.}
\end{subfigure}
\caption{\textbf{Global covariance and PSD for visual reading rollouts (60\,s context).}}
\label{fig:global_vis}
\end{figure*}

\begin{figure*}[!h]
\centering
\begin{subfigure}{0.49\linewidth}
\centering
\includegraphics[width=\linewidth]{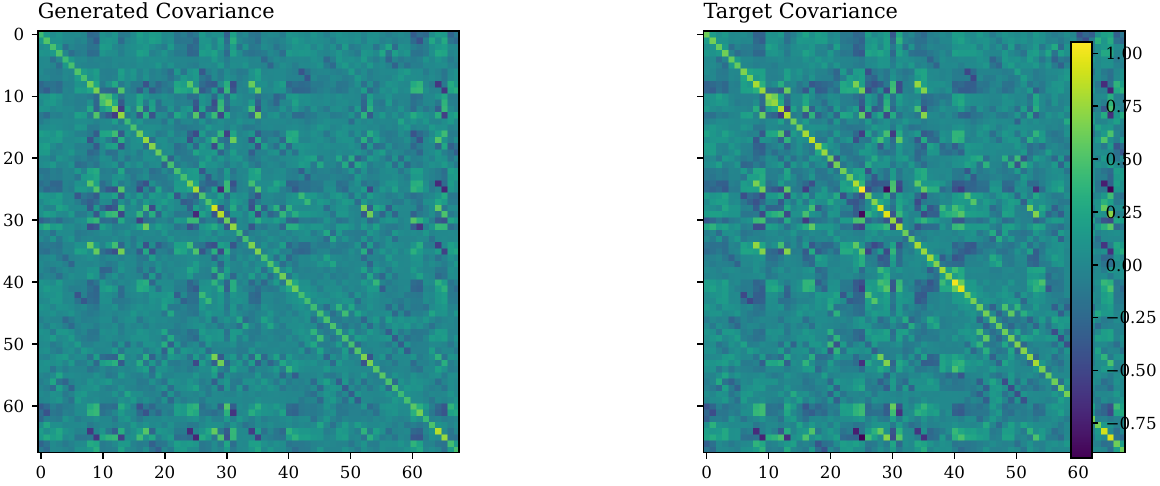}
\caption{Rest: covariance.}
\end{subfigure}
\hfill
\begin{subfigure}{0.49\linewidth}
\centering
\includegraphics[width=\linewidth]{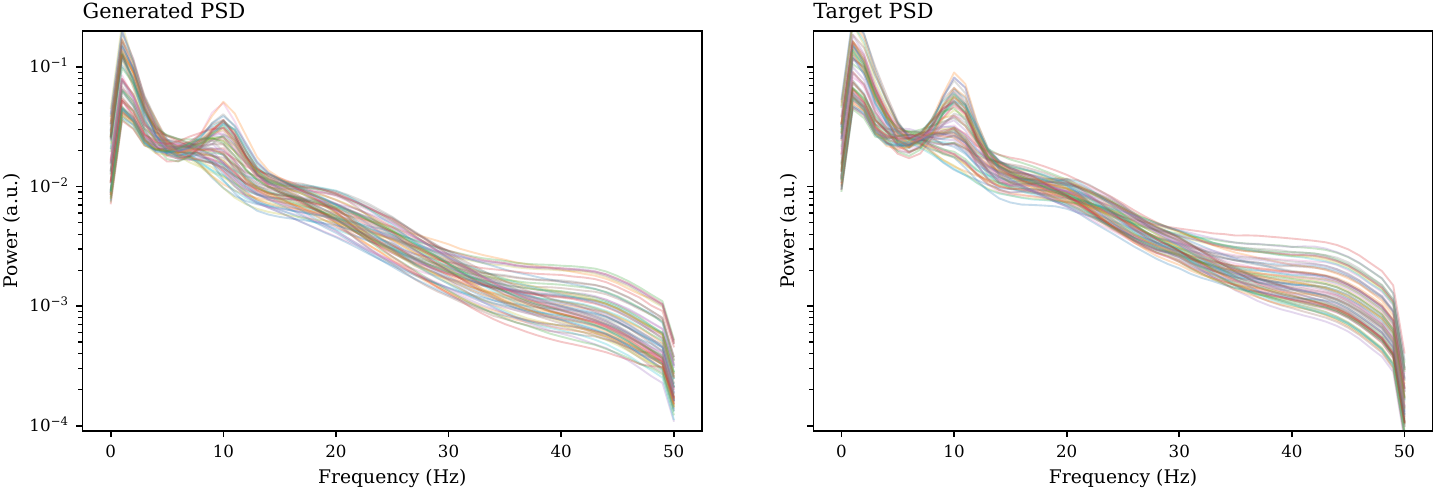}
\caption{Rest: PSD.}
\end{subfigure}
\caption{\textbf{Global covariance and PSD for resting-state rollouts (60\,s context).}}
\label{fig:global_rest}
\end{figure*}

\subsection{Full stability metrics for 60\,s-context rollouts}

\begin{figure}[!h]
\centering
\includegraphics[width=\linewidth]{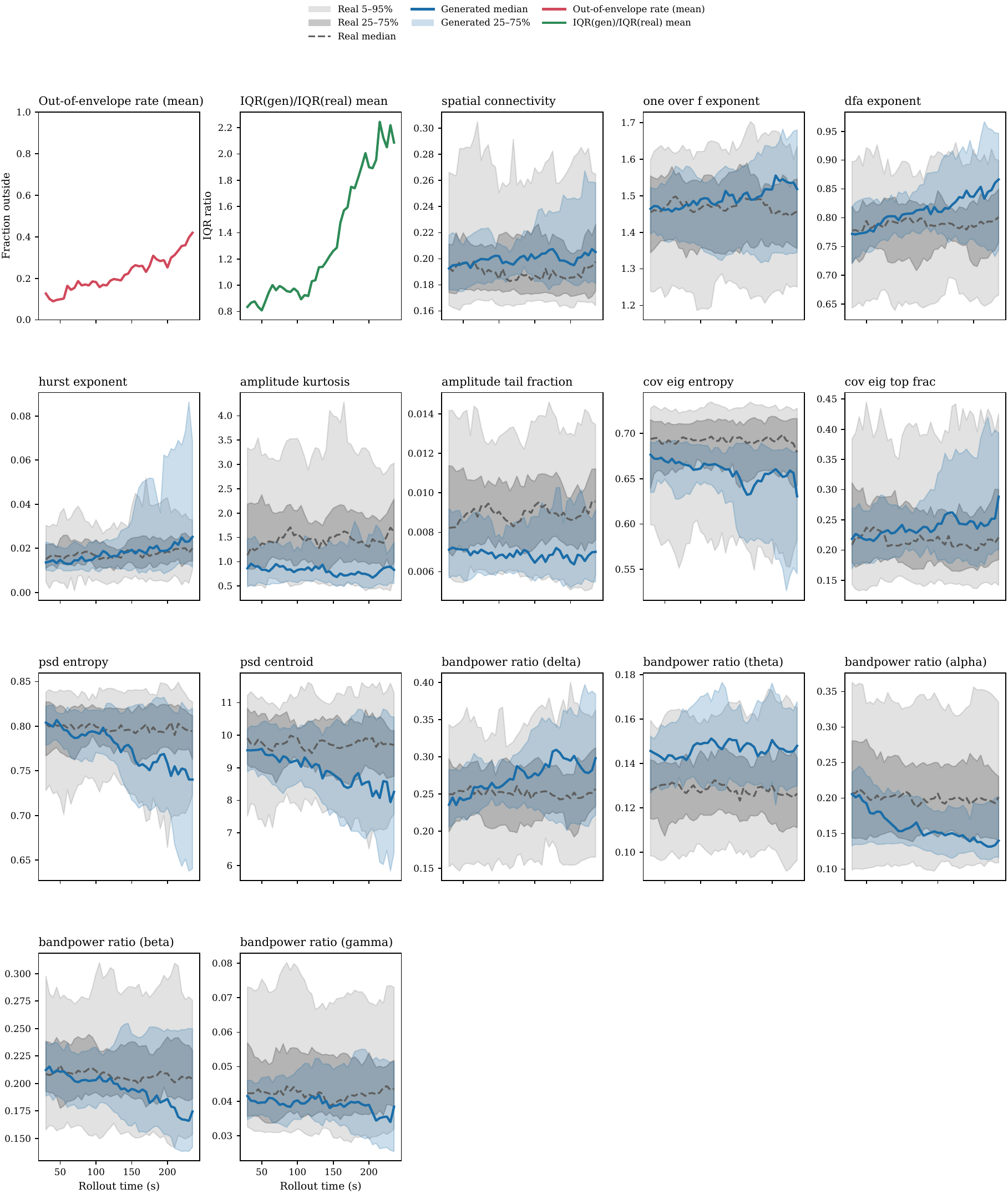}
\caption{\textbf{Auditory (60\,s context): full sliding-window stability.}}
\label{fig:window_full_aud}
\end{figure}

\begin{figure}[!h]
\centering
\includegraphics[width=\linewidth]{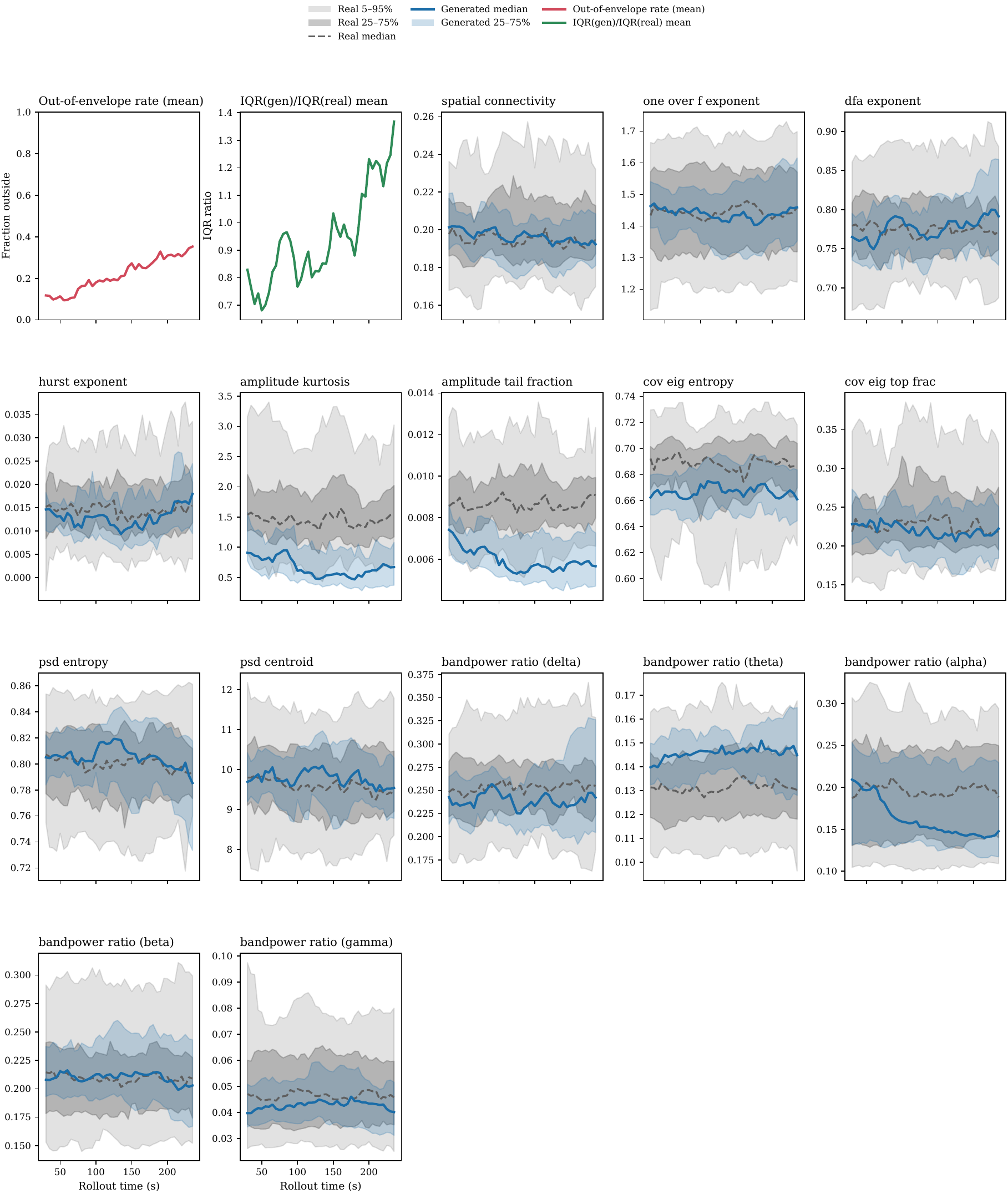}
\caption{\textbf{Visual (60\,s context): full sliding-window stability.}}
\label{fig:window_full_vis}
\end{figure}

\begin{figure}[!h]
\centering
\includegraphics[width=\linewidth]{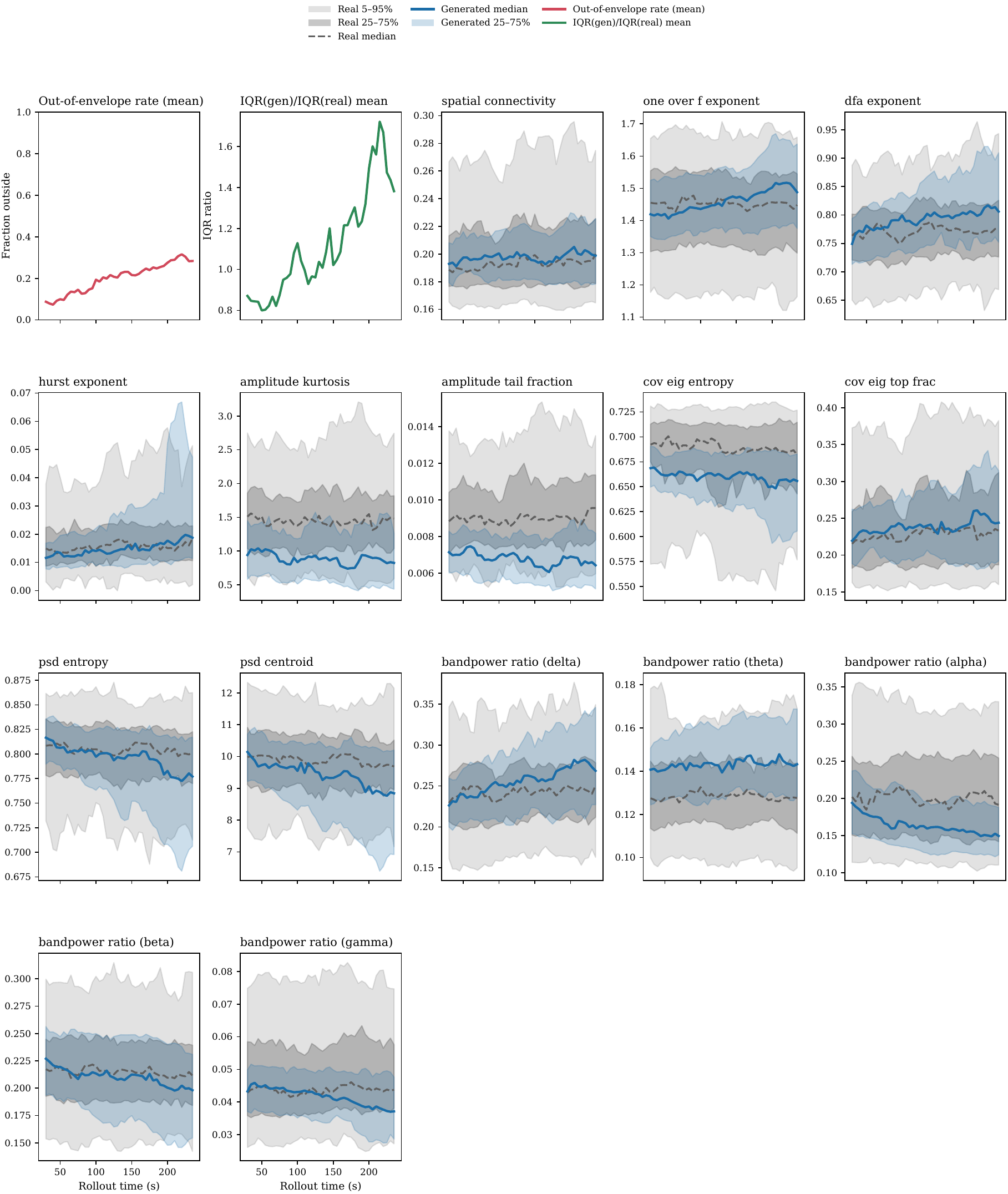}
\caption{\textbf{Rest (60\,s context): full sliding-window stability metrics.} Note that correlation and stft/fft angle are expect to have high distance due to phase/dynamics-misalignment between generated and real data.}
\label{fig:window_full_rest}
\end{figure}

\clearpage
\subsection{Full prefix-divergence metrics for 60\,s-context rollouts}

\begin{figure}[!h]
\centering
\includegraphics[width=\linewidth]{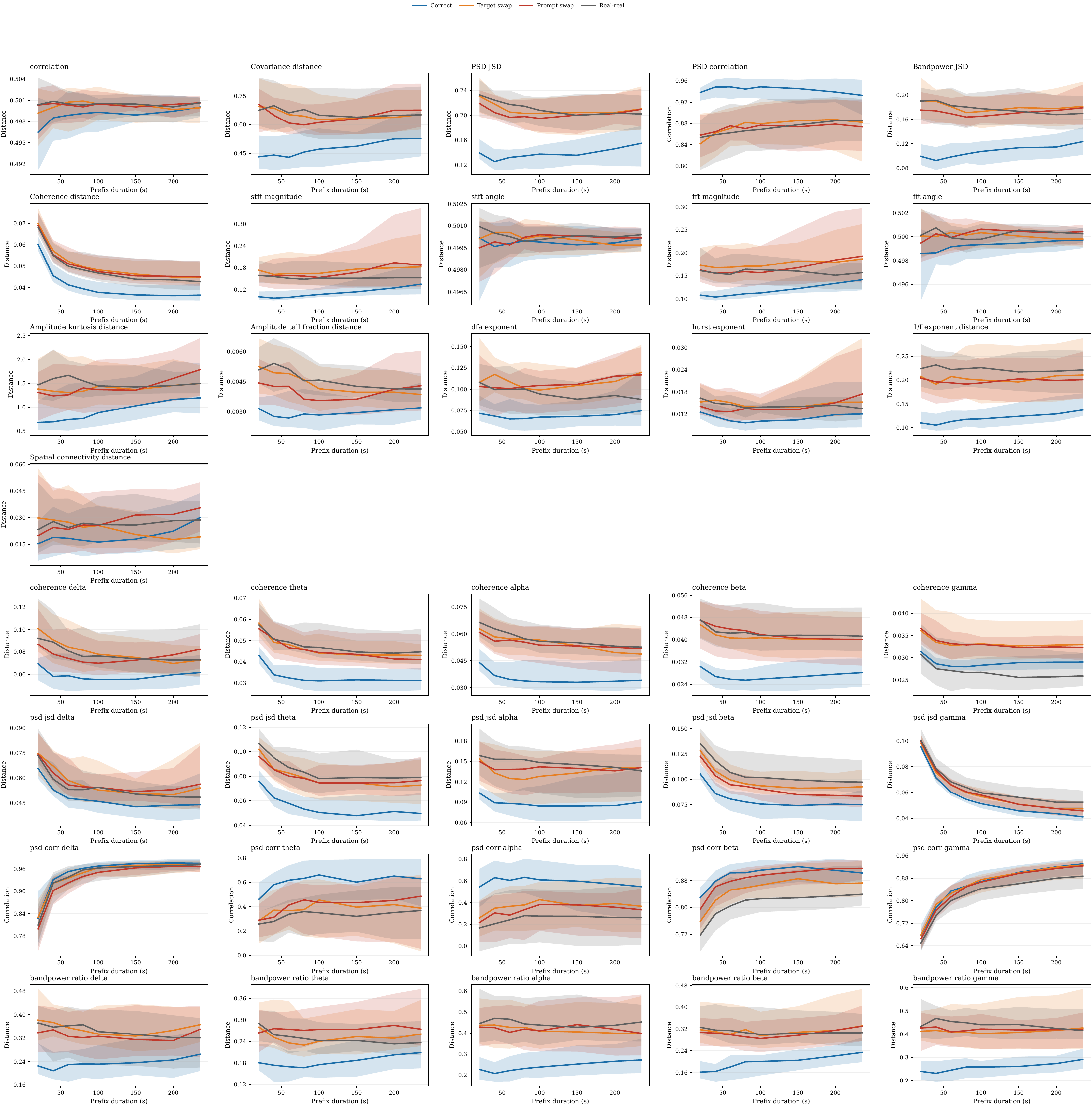}
\caption{\textbf{Auditory (60\,s context): full prefix-divergence metrics.} Note that correlation and stft/fft angle are expect to have high distance due to phase/dynamics-misalignment between generated and real data.}
\label{fig:div_full_aud}
\end{figure}

\begin{figure}[!h]
\centering
\includegraphics[width=\linewidth]{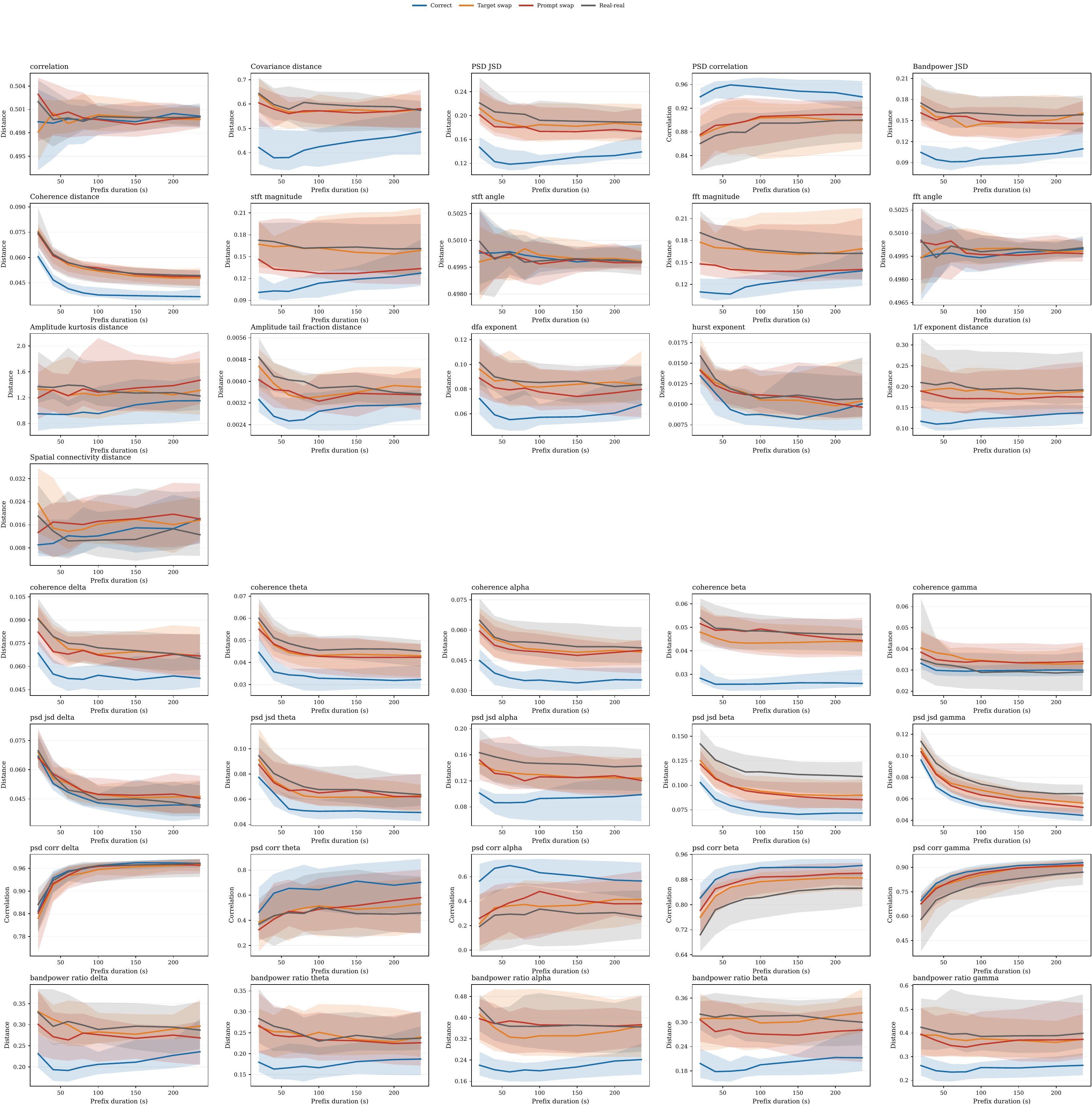}
\caption{\textbf{Visual (60\,s context): full prefix-divergence metrics.} Note that correlation and stft/fft angle are expect to have high distance due to phase/dynamics-misalignment between generated and real data.}
\label{fig:div_full_vis}
\end{figure}

\begin{figure}[!h]
\centering
\includegraphics[width=\linewidth]{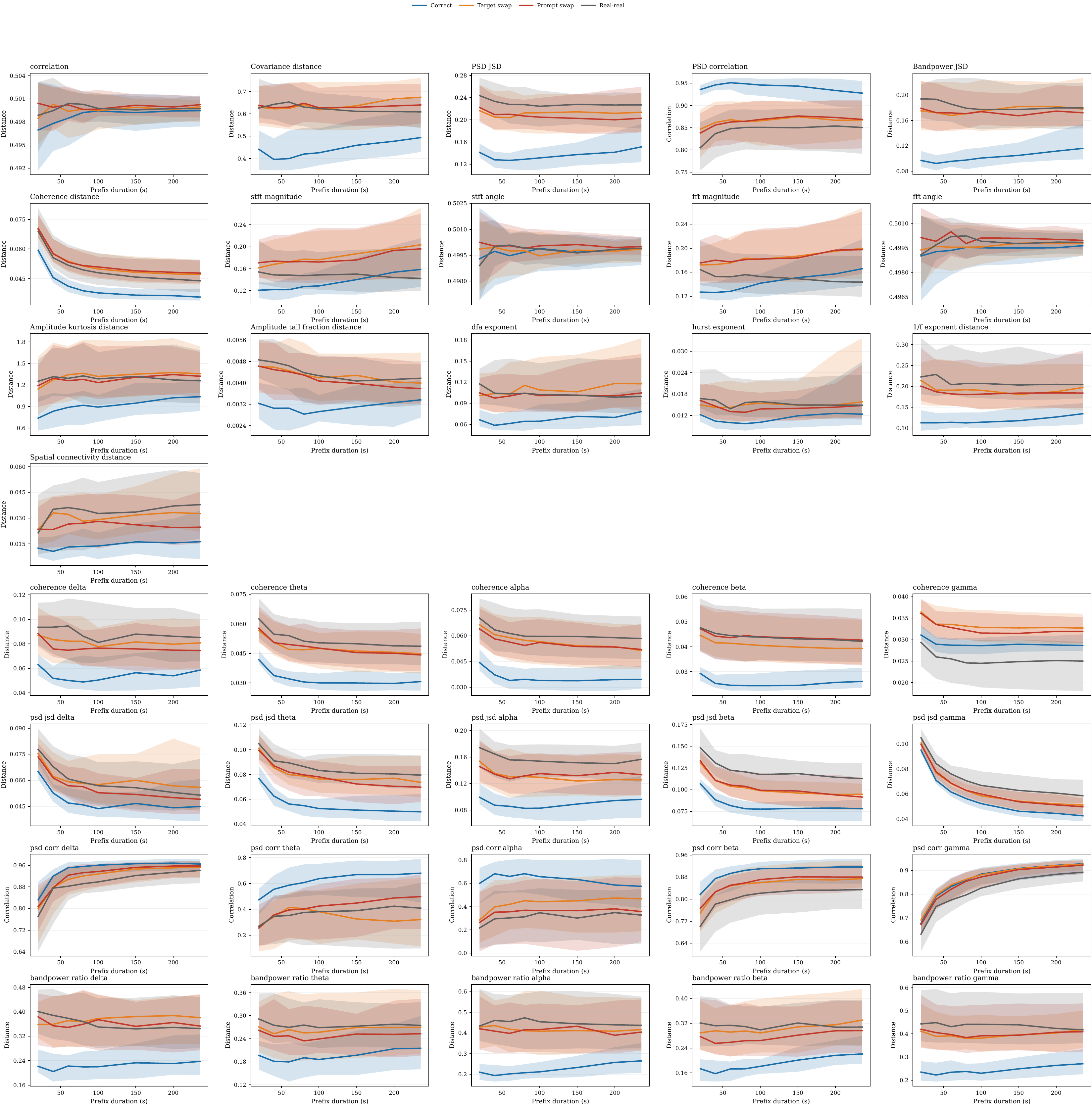}
\caption{\textbf{Rest (60\,s context): full prefix-divergence metrics.} Note that correlation and stft/fft angle are expect to have high distance due to phase/dynamics-misalignment between generated and real data.}
\label{fig:div_full_rest}
\end{figure}

\clearpage

\subsection{Qualitative rollouts}

\vspace{2.2cm}

\begin{figure}[!h]
\centering
\begin{subfigure}{0.48\textwidth}
\centering
\includegraphics[width=\linewidth]{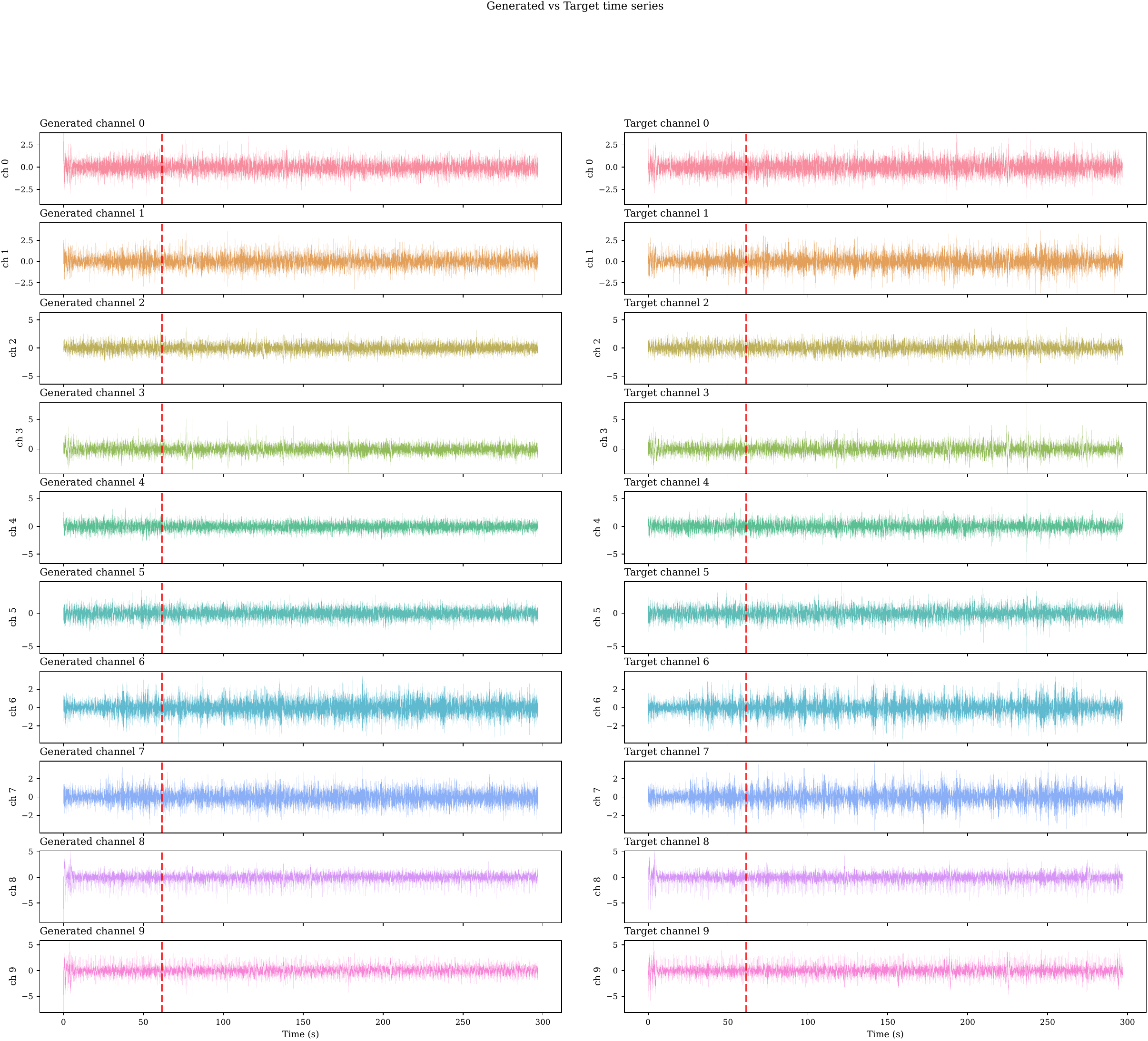}
\caption{Time series.}
\end{subfigure}
\hfill
\begin{subfigure}{0.48\textwidth}
\centering
\includegraphics[width=\linewidth]{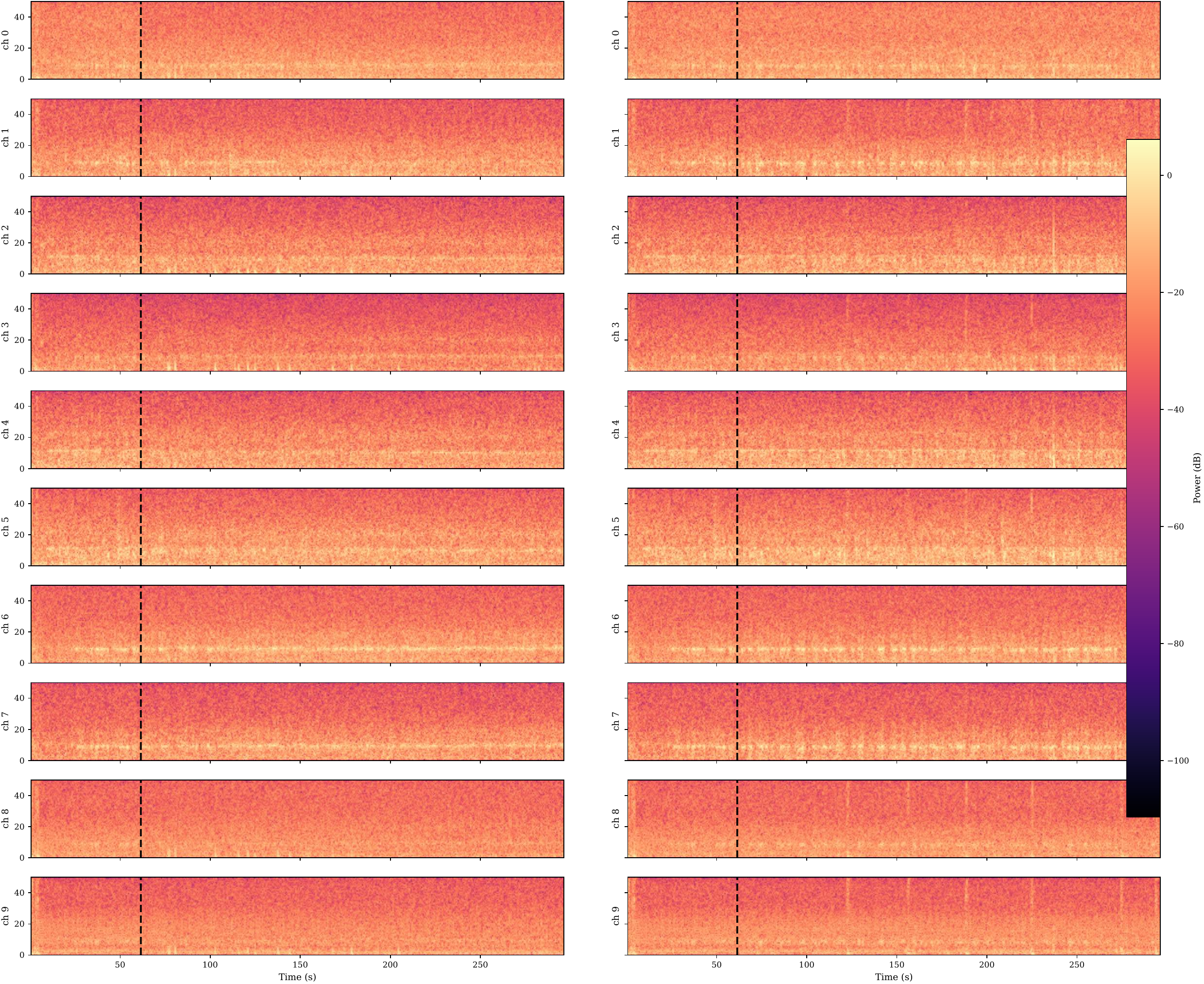}
\caption{STFT.}
\end{subfigure}
\caption{\textbf{Auditory qualitative rollout.} Dashed lines indicate boundary of context and continuation. 10 random channels are shown due to space constraints.}
\label{fig:qual_rollouts}
\end{figure}

\begin{figure}[!h]
\centering
\begin{subfigure}{0.48\textwidth}
\centering
\includegraphics[width=\linewidth]{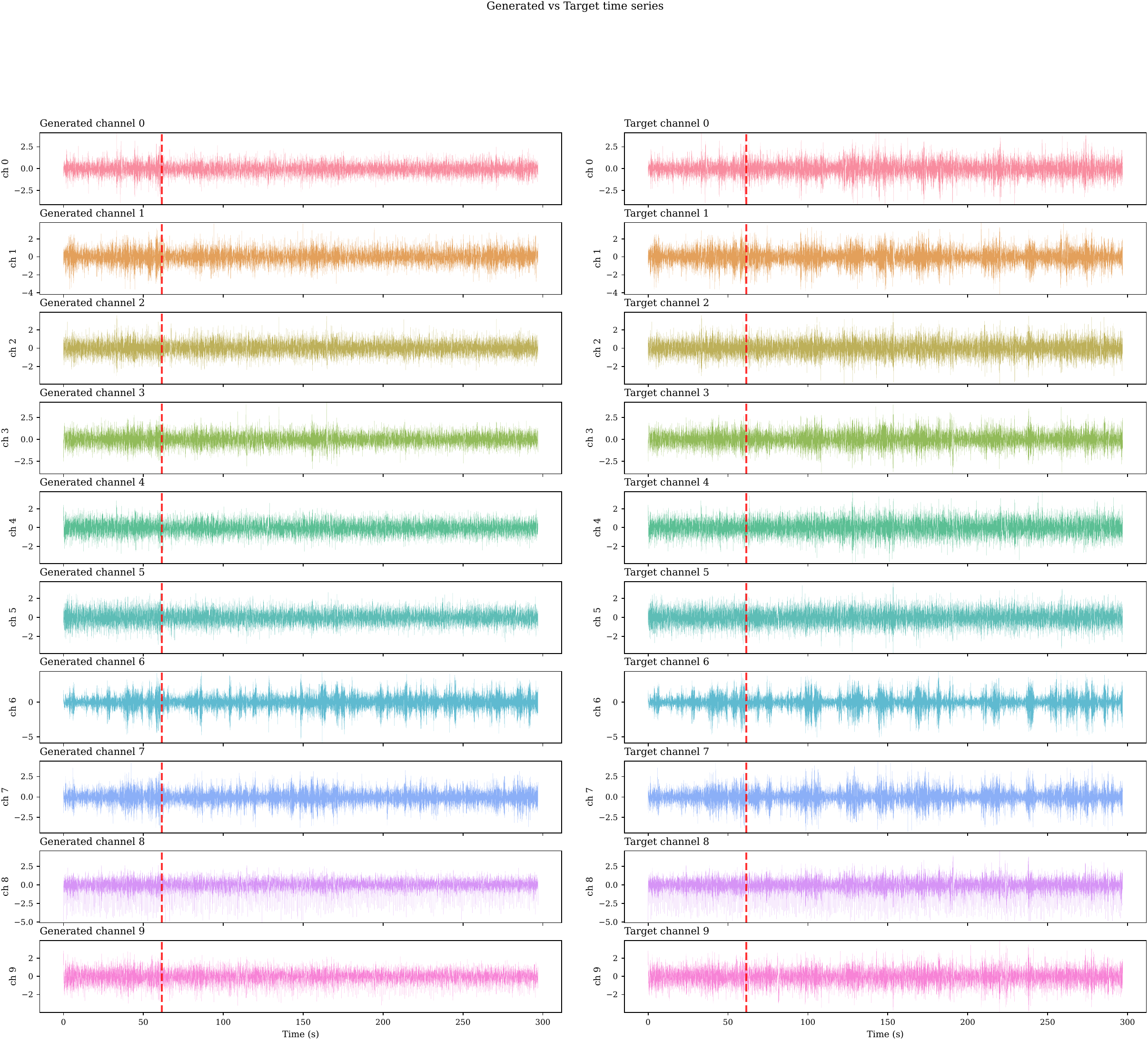}
\caption{Time series.}
\end{subfigure}
\hfill
\begin{subfigure}{0.48\textwidth}
\centering
\includegraphics[width=\linewidth]{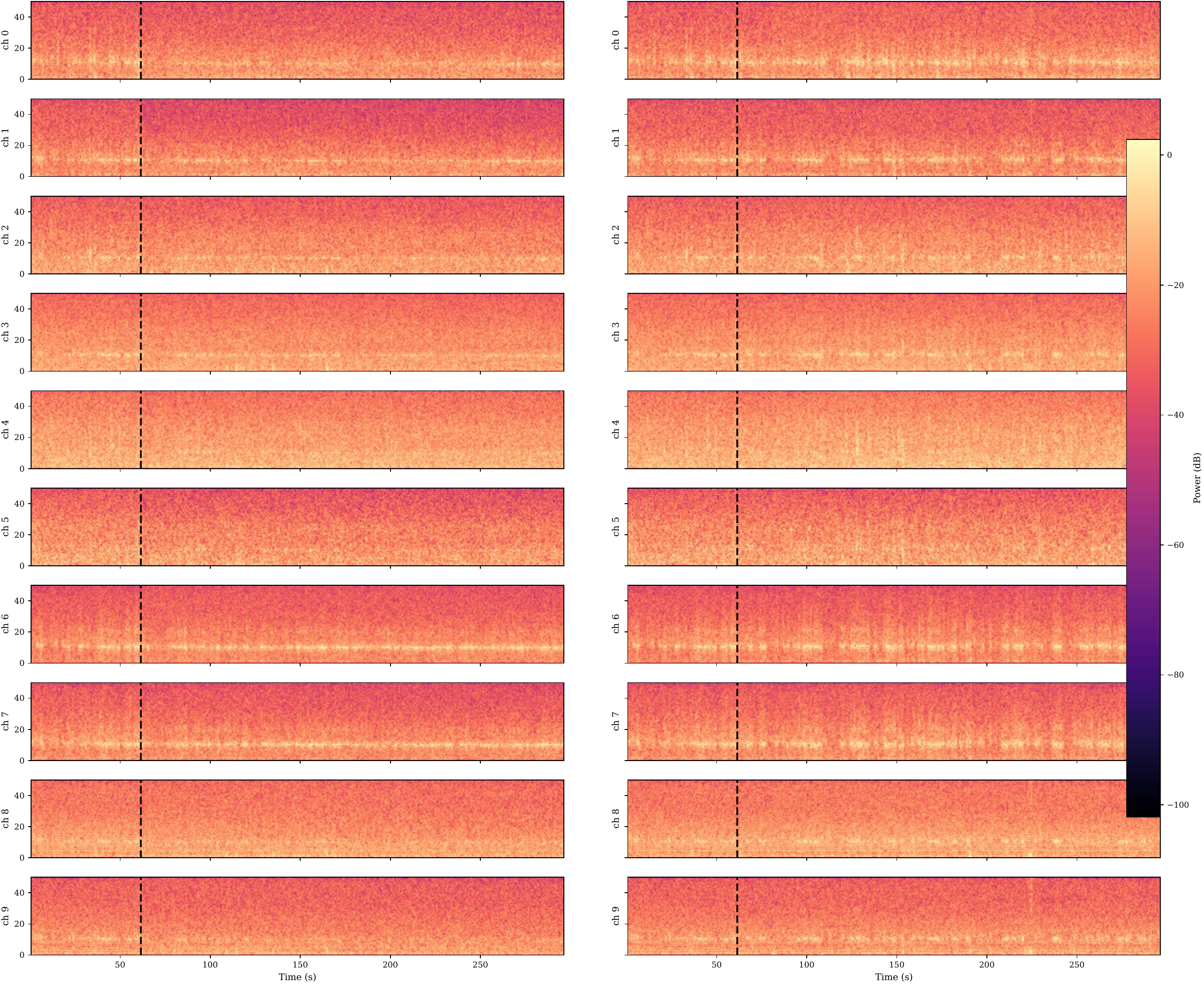}
\caption{STFT.}
\end{subfigure}
\caption{\textbf{Resting-state qualitative rollout.} Dashed lines indicate boundary of context and continuation. 10 random channels are shown due to space constraints.}
\label{fig:qual_rollouts_rest}
\end{figure}

\clearpage
\subsection{30\,s context ablation}
\label{sec:ablation_30s}

\begin{figure}[!h]
\centering
\includegraphics[width=\linewidth,height=0.85\textheight,keepaspectratio]{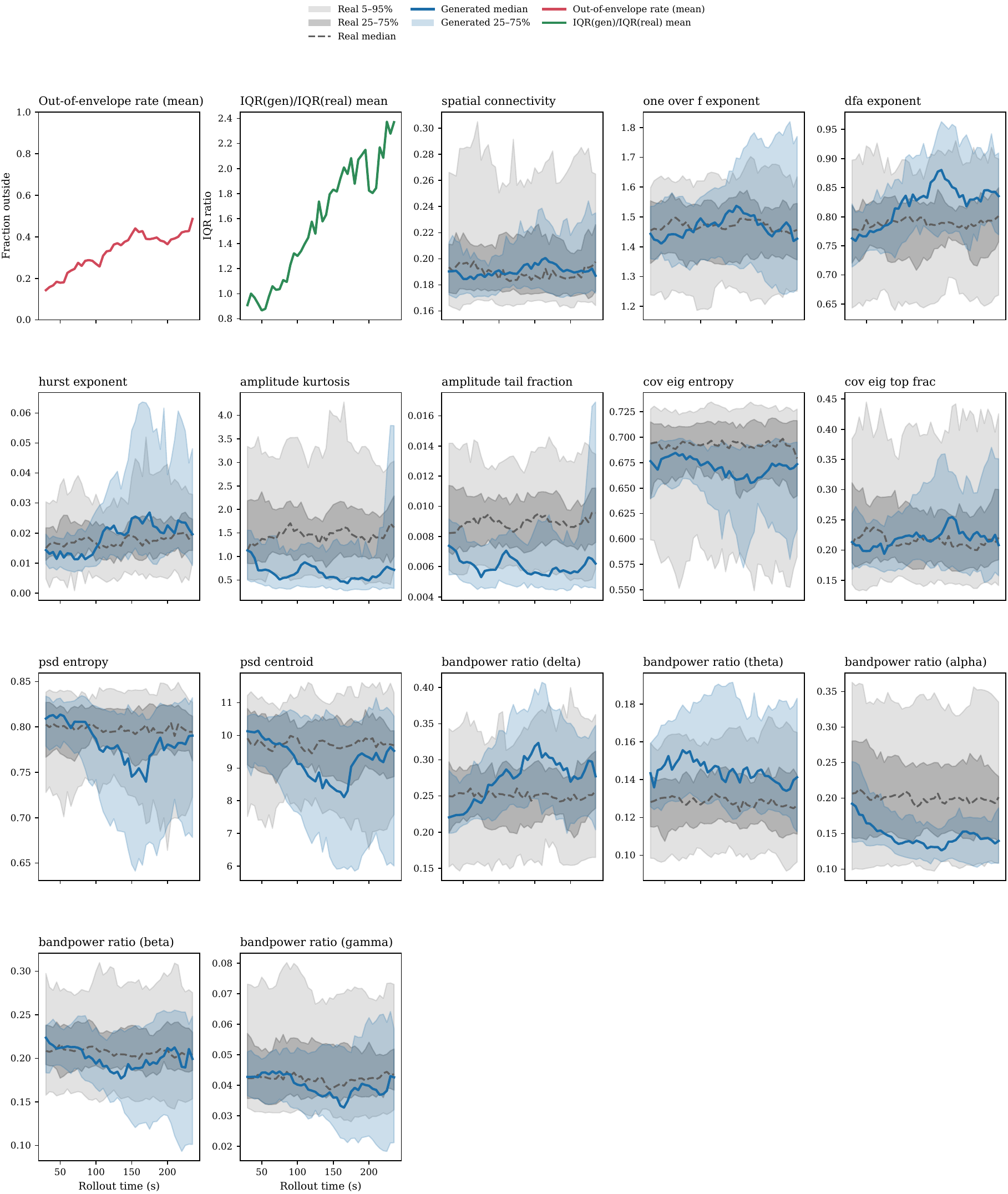}
\caption{\textbf{Auditory (30\,s context): full sliding-window stability.}}
\label{fig:ablation_30s_aud_window}
\end{figure}

\begin{figure}[!h]
\centering
\includegraphics[width=\linewidth,height=0.85\textheight,keepaspectratio]{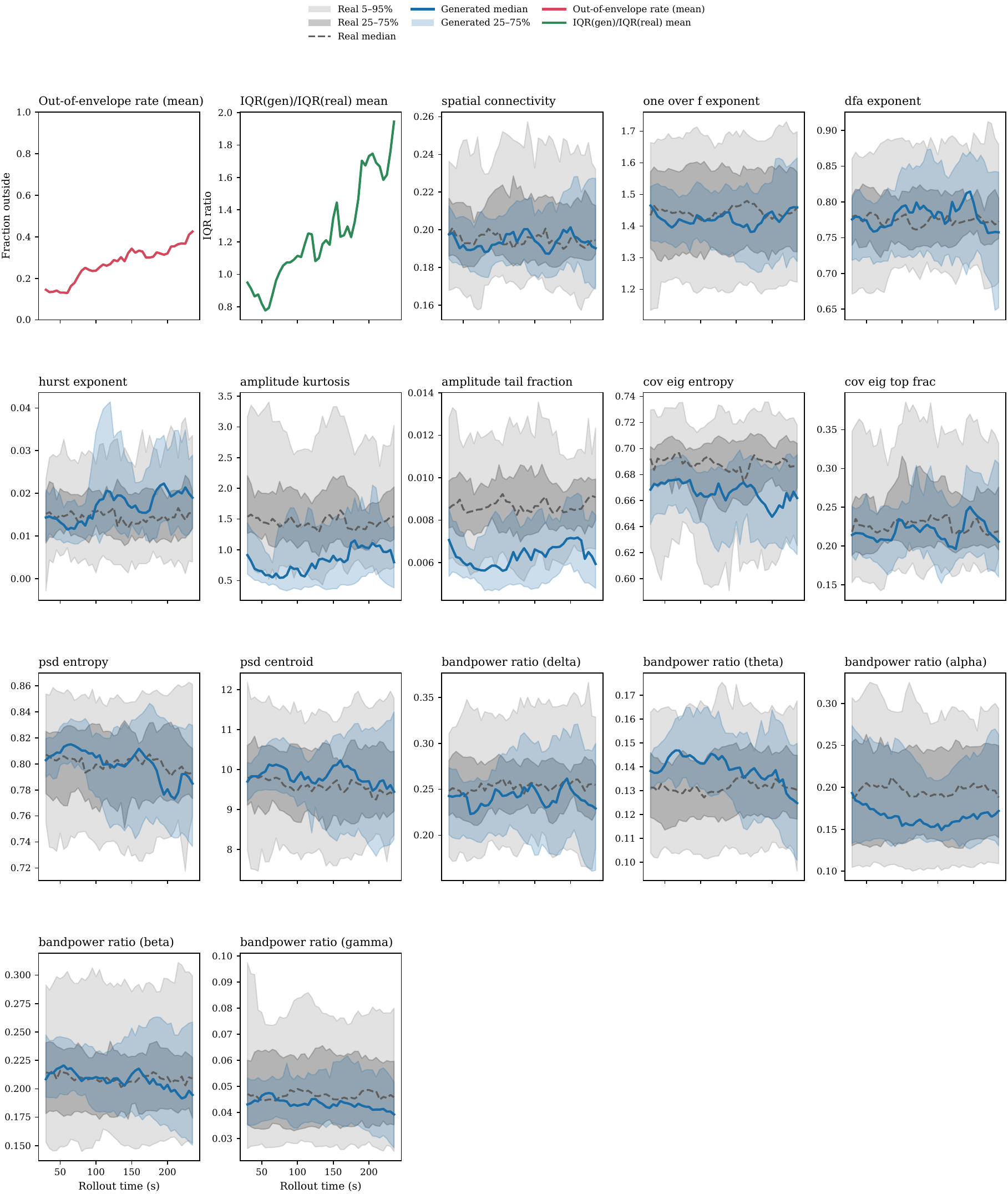}
\caption{\textbf{Visual (30\,s context): full sliding-window stability.}}
\label{fig:ablation_30s_window}
\end{figure}

\begin{figure}[!h]
\centering
\includegraphics[width=\linewidth,height=0.85\textheight,keepaspectratio]{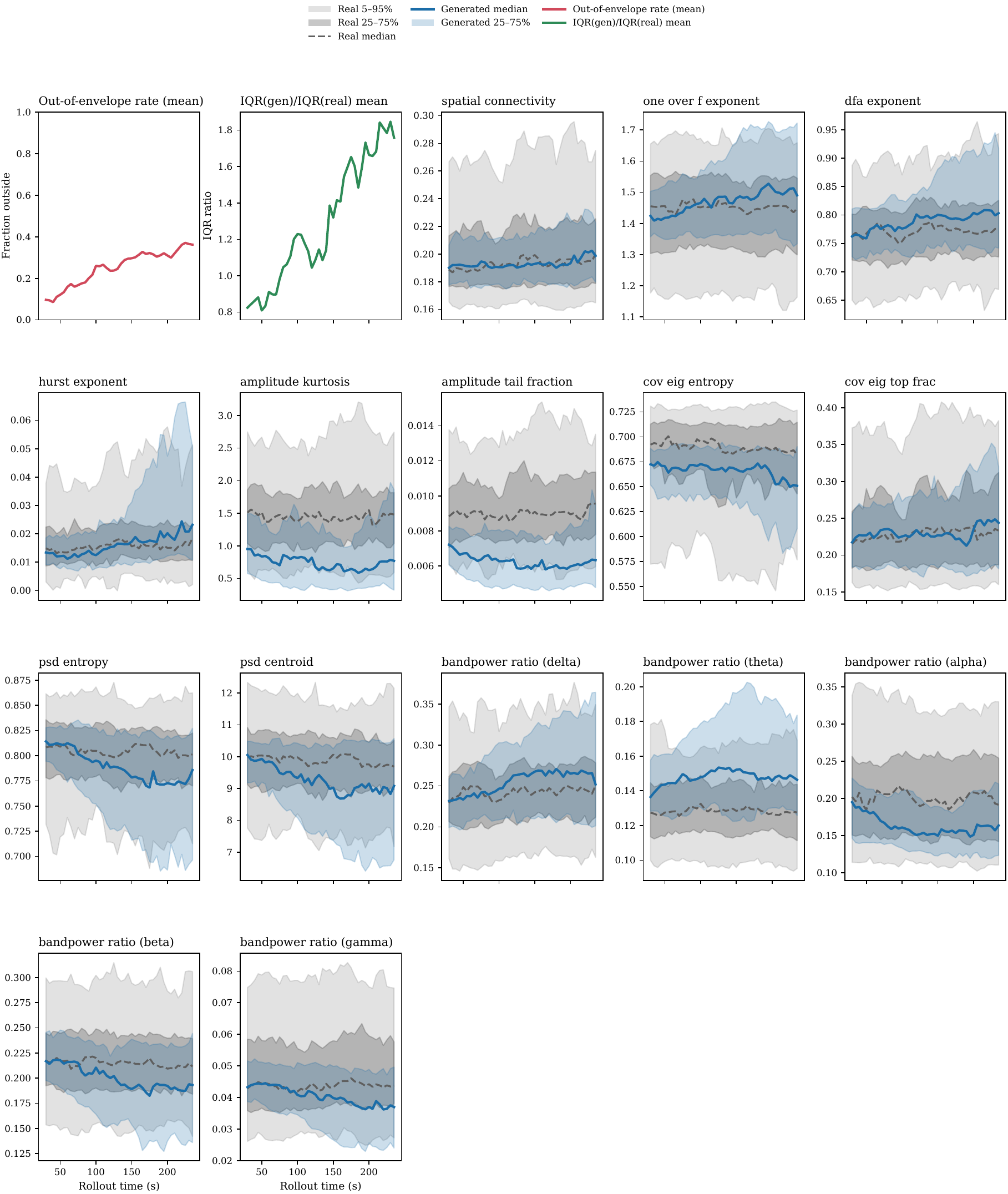}
\caption{\textbf{Rest (30\,s context): full sliding-window stability metrics.} Note that correlation and stft/fft angle are expect to have high distance due to phase/dynamics-misalignment between generated and real data.}
\label{fig:ablation_30s_rest_window}
\end{figure}

\clearpage

\begin{figure}[!h]
\centering
\includegraphics[width=\linewidth]{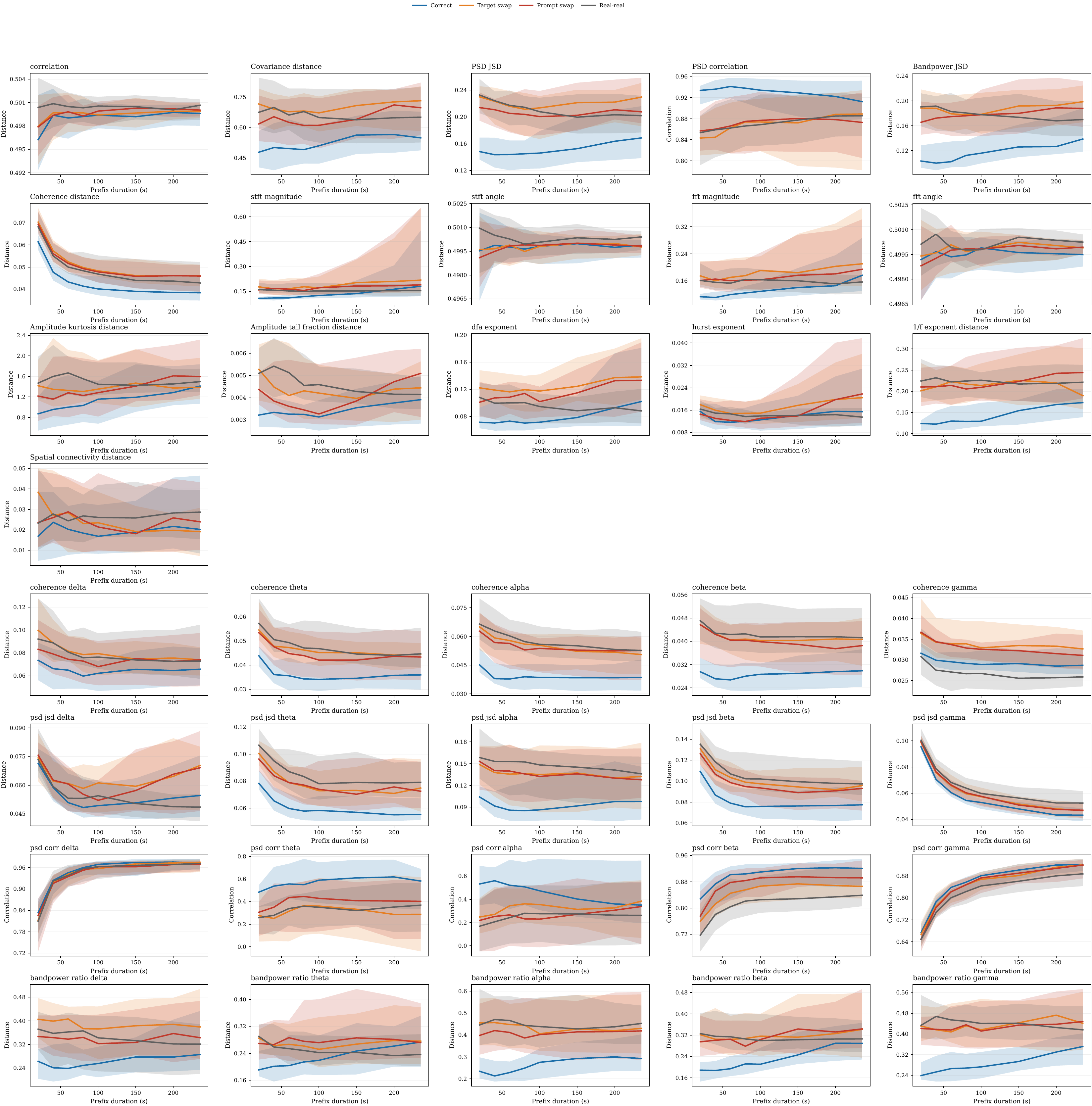}
\caption{\textbf{Auditory (30\,s context): full prefix-divergence metrics.} Note that correlation and stft/fft angle are expect to have high distance due to phase/dynamics-misalignment between generated and real data.}
\label{fig:ablation_30s_aud}
\end{figure}

\begin{figure}[!h]
\centering
\includegraphics[width=\linewidth]{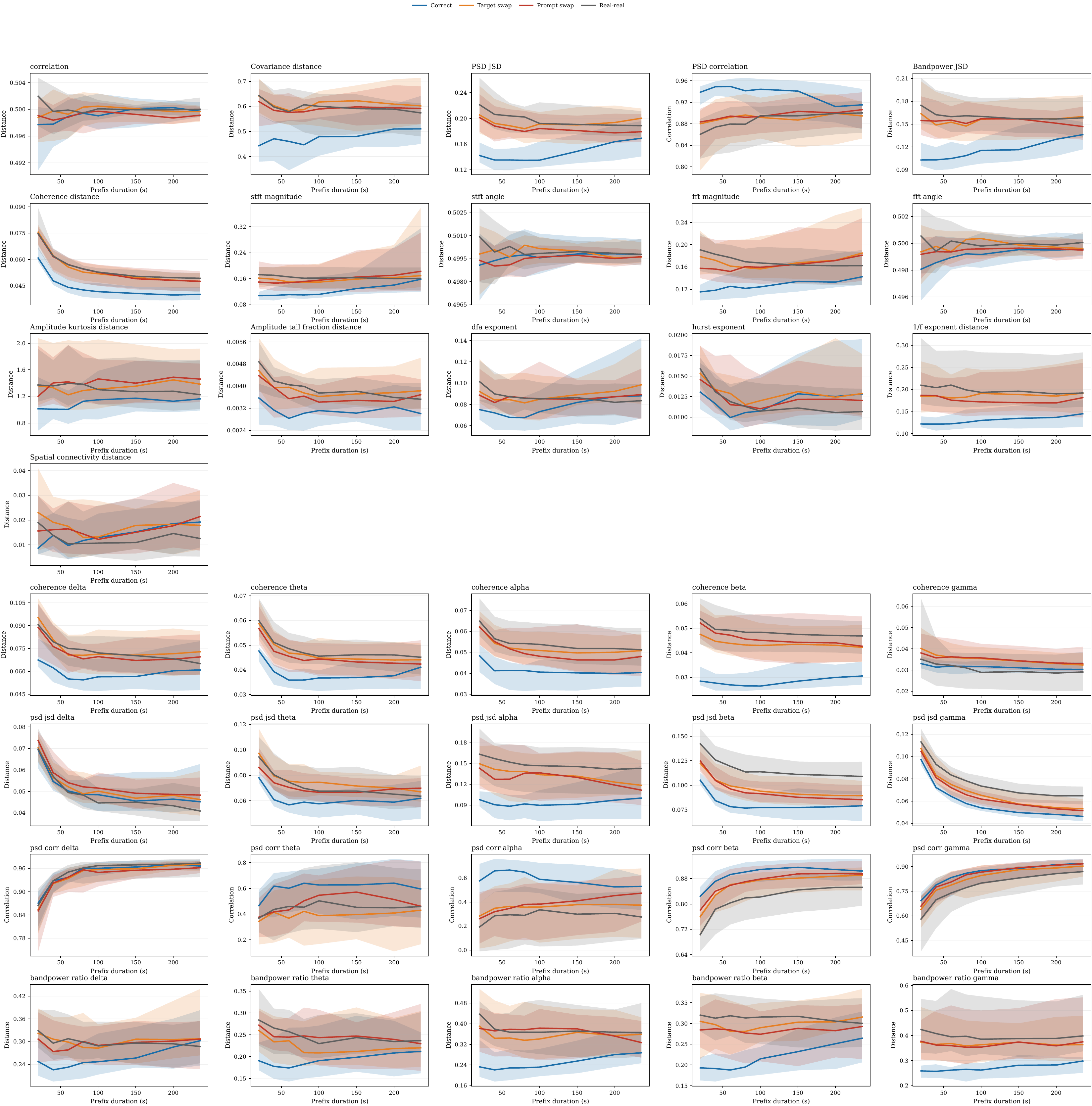}
\caption{\textbf{Visual (30\,s context): full prefix-divergence metrics.} Note that correlation and stft/fft angle are expect to have high distance due to phase/dynamics-misalignment between generated and real data.}
\label{fig:ablation_30s}
\end{figure}

\begin{figure}[!h]
\centering
\includegraphics[width=\linewidth]{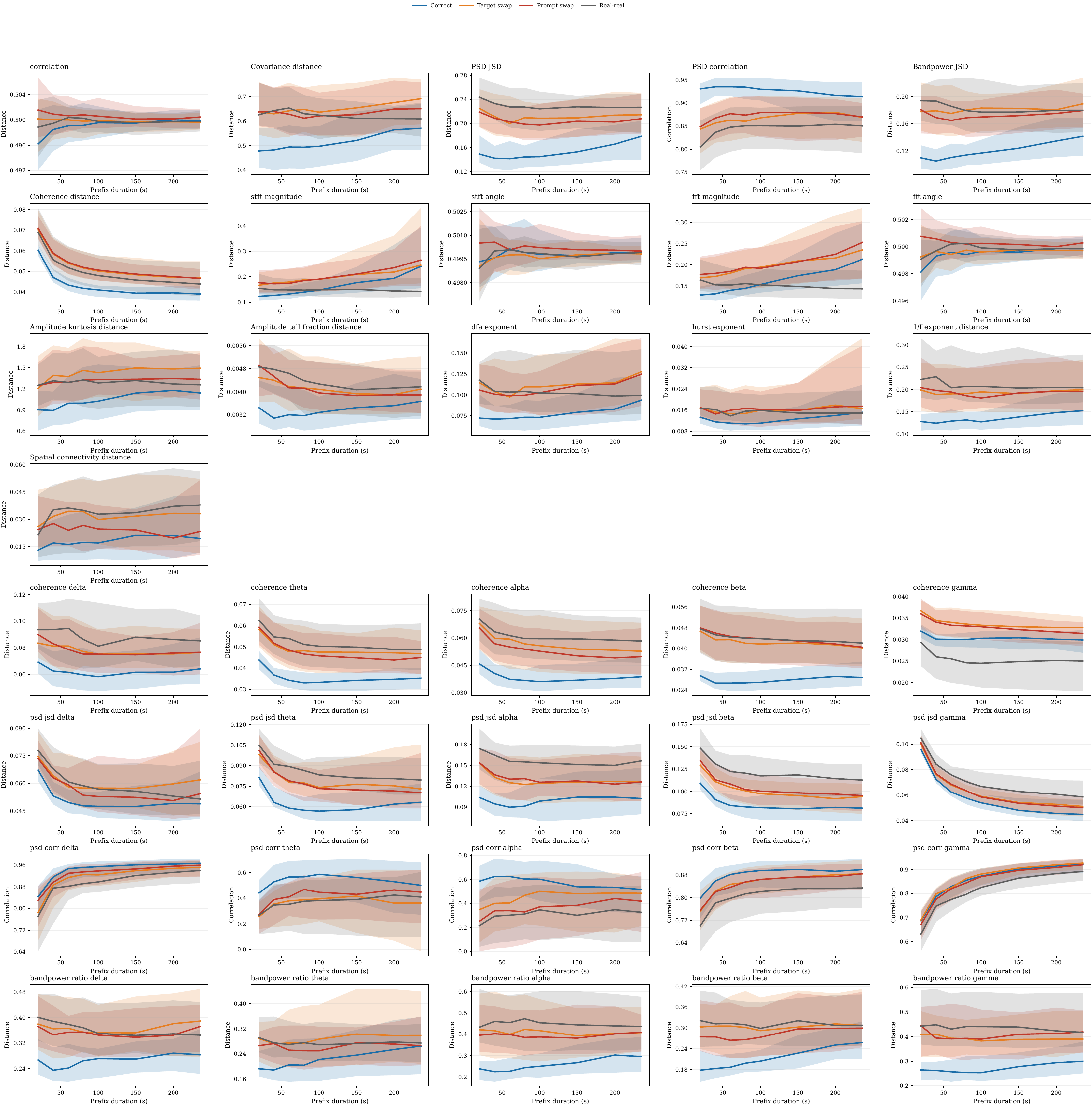}
\caption{\textbf{Rest (30\,s context): full prefix-divergence metrics.} Note that correlation and stft/fft angle are expect to have high distance due to phase/dynamics-misalignment between generated and real data.}
\label{fig:ablation_30s_rest}
\end{figure}

\end{document}